\documentclass[11pt]{article}
\usepackage[final]{acl}
\usepackage{graphicx}
\usepackage{dblfloatfix}
\usepackage{booktabs}
\usepackage{amsmath}
\usepackage{tikz}
\usepackage{pgf-pie}
\usepackage{fontspec}
\usepackage{xeCJK}
\usepackage{twemojis}
\usepackage{xcolor}
\usepackage{array}
\usepackage{colortbl}
\usepackage{multirow}
\usepackage{listings}
\usepackage[most]{tcolorbox}
\usetikzlibrary{arrows.meta,fit,positioning,shapes.geometric}

\setCJKmonofont[BoldFont=FandolSong-Bold.otf]{FandolSong-Regular.otf}

\definecolor{MeddiesPromptFrame}{RGB}{33, 112, 120}
\definecolor{MeddiesPromptBack}{RGB}{248, 252, 252}
\definecolor{MeddiesPromptTitle}{RGB}{227, 242, 243}
\definecolor{MeddiesChartPrimary}{RGB}{86, 180, 233}
\definecolor{MeddiesChartSecondary}{RGB}{230, 159, 0}
\definecolor{MeddiesChartNeutral}{RGB}{0, 158, 115}

\lstdefinestyle{meddiesprompt}{
  basicstyle=\ttfamily\scriptsize,
  breaklines=true,
  breakatwhitespace=false,
  breakindent=0pt,
  breakautoindent=false,
  columns=fullflexible,
  keepspaces=true,
  showstringspaces=false
}



\begin{document}

\title{Meddies-PII: A Multilingual Framework for Personally Identifiable Information Extraction in Clinical De-identification}

\author{
  Le Linh Uyen\textsuperscript{*} \and
  Christian Hoang \and
  Ho Dinh Tri \and
  Huy Hoang Ha\textsuperscript{*} \\
  Meddies AI, Vietnam \\
  \texttt{hoangha@meddies.ai} \\
  \textsuperscript{*}These authors contributed equally
}

\maketitle

\hypersetup{
  pdftitle={Meddies-PII: A Multilingual Framework for Personally Identifiable Information Extraction in Clinical De-identification},
  pdfauthor={Le Linh Uyen; Christian Hoang; Ho Dinh Tri; Huy Hoang Ha},
  pdfsubject={Multilingual clinical PII extraction using synthetic data}
}

\begin{abstract}
Multilingual PII extraction in clinical text is limited by scarce governed data and by synthetic corpora whose construction is thinly documented. We present a large multilingual dataset covering seventeen languages, including seven Southeast Asian languages, and nine PII labels. We also document a detailed generation pipeline and release the complete prompts, catalogues, checks and label maps. A compact 350M-parameter BIOES token classifier trained on the resource achieves the highest score on all fifteen public benchmarks, significantly ahead of the strongest of four public systems on fourteen of them, with a mean exact-match F1 of 0.827 against 0.672 for the strongest system. We also tested it, without adaptation, on multilingual benchmarks, including two human-annotated German and Spanish clinical-style corpora, where it is competitive with the best public systems once annotation conventions are aligned, and is best in four of ten MultiGraSCCo languages; the results show that performance transfers across public benchmarks. We also analyse the data and find almost no near-duplicates and no text overlap between training and any evaluation set. The dataset, benchmark, model, generation framework and evaluation code are planned for public release upon acceptance.
\end{abstract}

\noindent\textbf{Keywords:} Natural Language Processing; Named Entity Recognition; Personally Identifiable Information; Synthetic Data; Healthcare\par\medskip

\begin{table*}[!t]
\centering
\small
\caption{Corpora relevant to this work. Statistics of external corpora are taken from the files we used or from the cited sources; ``--'' indicates that the value was not established. Formats gives the number of distinct text formats and Labels the number of annotated labels. Further details on our benchmark are in Appendix~\ref{app:benchmark}.}\label{tab:corpora}
\setlength{\tabcolsep}{3.5pt}
\begin{tabular}{>{\raggedright\arraybackslash}p{0.17\textwidth} >{\raggedright\arraybackslash}p{0.15\textwidth} >{\raggedright\arraybackslash}p{0.05\textwidth} >{\raggedleft\arraybackslash}p{0.11\textwidth} >{\raggedright\arraybackslash}p{0.10\textwidth} >{\raggedright\arraybackslash}p{0.09\textwidth} >{\raggedright\arraybackslash}p{0.05\textwidth} >{\raggedright\arraybackslash}p{0.16\textwidth}}
\toprule
Corpus & Text origin & Lang. & Documents & Annotation & Formats & Labels & Controlled variables \\
\midrule
i2b2/UTHealth 2014 \cite{stubbs2015automated} & real clinical notes & en & 1,304 & manual & -- & -- & -- \\
\arrayrulecolor{black!25}\midrule[0.3pt]\arrayrulecolor{black}
MEDDOCAN \cite{marimon2019meddocan} & clinical case reports, surrogate PHI & es & 1,000 & manual & 1 & 7 & -- \\
\arrayrulecolor{black!25}\midrule[0.3pt]\arrayrulecolor{black}
GraSCCo\_PHI \cite{lohr2024deidentifying} & alienated clinical documents & de & 63 & manual & 1 & 7 & -- \\
\arrayrulecolor{black!25}\midrule[0.3pt]\arrayrulecolor{black}
OpenPII \cite{ai4privacy_openpii_1_5m_2026} & synthetic & 12 & 1,636,375 & Unknown & 1 & 19 & language, region \\
\arrayrulecolor{black!25}\midrule[0.3pt]\arrayrulecolor{black}
Gretel PII \cite{gretel-pii-docs-en-v1} & LLM-generated & en & $\approx$60k & at generation, auto-validated & -- & 42 & domain, document type \\
\arrayrulecolor{black!25}\midrule[0.3pt]\arrayrulecolor{black}
Nemotron-PII \cite{nemotron-pii} & synthetic personas & en & 100,000 & at generation & 2 & 55+ & domain, document type, locale, persona \\
\arrayrulecolor{black!45}\midrule[1.2pt]\arrayrulecolor{black}
\textbf{Our dataset} & \textbf{LLM-generated, clinical-style} & \textbf{17} & \textbf{799,306} & \textbf{LLM tags + deterministic checks} & \textbf{20} & \textbf{9} & \textbf{attribute-conditioned prompting} \\
\bottomrule
\end{tabular}
\end{table*}

\section{Introduction}

Clinical notes and other privacy-sensitive text can be shared and analysed only after personally identifiable information (PII) has been found and removed \cite{savkin-etal-2025-spy}. PII extraction is usually cast as named entity recognition, but its ontology and error costs differ from general-purpose NER: a missed identifier is far more harmful than a spurious one. Manually annotated de-identification corpora are expensive to build and therefore cover few languages and domains \cite{neamatullah2008automated,uzuner2007evaluating,stubbs2015automated}. Several large synthetic PII datasets exist, but most do not fully disclose how they were produced or validated \cite{ai4privacy_openpii_1_5m_2026,gretel-pii-docs-en-v1,nemotron-pii}.

Therefore, multilingual training data whose construction can be inspected and reproduced remains a gap in the literature. We ask whether a controlled, fully documented generation pipeline with deterministic validation can fill it when governed clinical text is unavailable. Our dataset combines scale with breadth, covering seventeen languages, twelve document types, eleven text formats, and both structured and adversarial surface forms, and with reproducibility, since every prompt, catalogue and check is published. The documents are clinically styled (discharge summaries, referral letters, HL7 and FHIR messages, portal logs), generated by LLMs, and tagged by the same LLMs.

In addition, we trained a compact 350M-parameter model on the resource. On fifteen public benchmarks it averages 0.827 exact-match F1 against 0.672 for the strongest of four public systems; it scores highest on all fifteen and is significantly ahead of the strongest baseline on fourteen, with no significant difference on CredData \cite{sr-cred21} (95\% intervals in Appendix~\ref{app:ci}). This paper makes three contributions:
\begin{enumerate}
\item \textbf{A large multilingual resource.} Our dataset is a large seventeen-language, nine-label table of 799,306 generated clinical-style rows. We state its exact composition (Section~\ref{sec:composition}) and analyse its redundancy and overlap with the evaluation sets (Section~\ref{sec:audit}).
\item \textbf{A fully documented generation pipeline.} We publish the complete prompts, the document, format and scenario catalogues with descriptions, the thirteen surface perturbations, every verification check, and the label maps used for scoring.
\item \textbf{A compact model and a broad evaluation.} A 350M-parameter token classifier trained on the resource obtains the highest score against four public systems on all fifteen public benchmarks, significantly ahead on fourteen, and is competitive on human-annotated datasets (Section~\ref{sec:eval}).
\end{enumerate}

\section{Related Work}

\paragraph{Manually annotated corpora.} Annotated de-identification resources support supervised training and evaluation, but annotation cost limits their scale and language coverage \cite{deleger2014preparing}. Early English work includes a system of lookup tables, regular expressions and heuristics evaluated on nursing notes \cite{neamatullah2008automated} and the i2b2 challenges: one on discharge summaries with synthesized surrogate PHI, where the best systems scored above 98\% F-measure on every PHI category but ambiguous PHI remained hard \cite{uzuner2007evaluating}, and the 2014 edition on longitudinal narratives with a broader PHI set than HIPAA, where three of ten systems exceeded F1 of 0.90 \cite{stubbs2015automated}. Non-English resources include French clinical notes with ten PHI categories \cite{grouin2014identification}; the Spanish MEDDOCAN track, built on a synthetic corpus of clinical case documents, whose top systems used deep learning \cite{marimon2019meddocan}; the German GRASCCO, about 60 clinical documents obfuscated through multiple alienation steps so that they can be shared without legal restrictions \cite{modersohn2022grascco}, with PHI annotations at an inter-annotator agreement of Krippendorff's $\alpha \approx 0.97$ \cite{lohr2024deidentifying}; its translation into ten languages \cite{baroud2026multigrascco}; and four Spanish legal-domain datasets, one annotated automatically with an estimated error rate of about 14\% \cite{de-gibert-bonet-etal-2022-spanish}. Table~\ref{tab:corpora} compares the main PII corpora with our dataset.

\paragraph{Synthetic corpora.} Synthetic generation scales to many languages: OpenPII 1.5M has 1,636,375 examples in 30 languages with 19 labels \cite{ai4privacy_openpii_1_5m_2026}, Nemotron-PII has 100,000 English persona-grounded records across more than 50 industries \cite{nemotron-pii}, and Gretel's English documents were generated with Gretel Navigator for fine-tuning GLiNER models \cite{gretel-pii-docs-en-v1}. The controls used for diversity, label coverage and validation are not always specified. SPY documents its workflow more fully, using LLMs to generate a PII detection dataset on which dedicated NER models are limited \cite{savkin-etal-2025-spy}. Attribute-conditioned prompting is a standard way to raise the diversity of generated data: prompts that specify attributes such as length and style outperform simple class-conditional prompts, which yield biased data \cite{yu2023large}. Our dataset uses it with complete documentation for multilingual, structured, clinical-style PII.

\paragraph{PII models.} PII extraction specializes sequence labeling: transformer encoders classify tokens, optionally with a CRF or constrained decoding \cite{souza2019portuguese,devlin2019bert,lester-etal-2020-constrained}, the latter suppressing illegal tag transitions and training about twice as fast as a CRF with statistically insignificant differences in F1 \cite{lester-etal-2020-constrained}. Span-based and schema-driven generalist models such as GLiNER2 predict entities of requested types \cite{eberts2019span,zaratiana2023gliner,zaratiana2025gliner2efficientmultitaskinformation}; GLiNER uses a compact bidirectional encoder and outperforms ChatGPT and fine-tuned LLMs in zero-shot NER \cite{zaratiana2023gliner}. OpenMed-PII is a DeBERTa-v3-large model trained on Nemotron-PII for 54 types, English only according to its model card \cite{openmed-pii-2026}; the LFM2.5 detector is a full fine-tune of LFM2.5-Encoder-350M for 40 types in 16 languages \cite{liquidai2026lfm25pii}; and OpenAI Privacy Filter is a dedicated PII system \cite{openai2026privacyfilter}.

\section{Our Dataset}

\subsection{Overview}
\label{sec:composition}

Our dataset consists of the 799,306 rows generated by our pipeline, in two collections: an initial one of 572,760 rows (71.7\%) and a refined one of 226,546 rows (28.3\%) that was created after we reviewed the first and revised the generation procedure \cite{xi-etal-2025-samplemix}. Every row is a synthetic clinical-style document with character-level annotations of nine labels, drawn from 12 document types, 11 text formats and 12 scenarios, and written by one of nine generator LLMs (Section~\ref{sec:pipeline}).

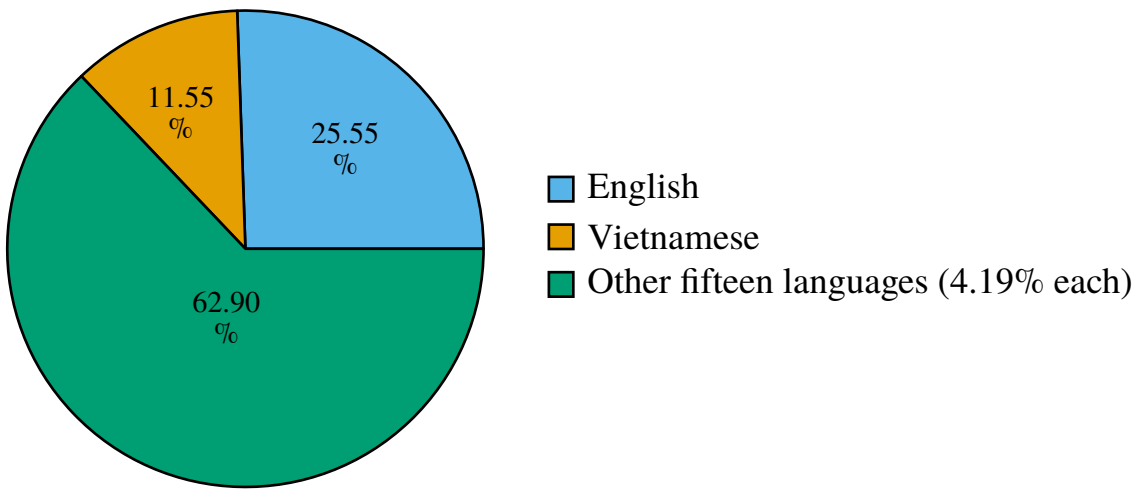
\begin{figure}[!htbp]
\centering
\resizebox{\columnwidth}{!}{%
\begin{tikzpicture}
\pie[
    sum=100,
    text=legend,
    hide number=true,
    radius=2.50,
    color={MeddiesChartPrimary,MeddiesChartSecondary,MeddiesChartNeutral}
]{25.55/English,11.55/Vietnamese,62.90/Other fifteen languages (4.19\% each)}
\node at (1.05,1.05) {\shortstack{\small 25.55 \\ \small\% }};
\node at (-0.68,1.43) {\shortstack{\small 11.55 \\ \small\% }};
\node at (-0.20,-0.72) {\shortstack{\small 62.90 \\ \small\% }};
\end{tikzpicture}%
}
\caption{Language distribution of the one-million-row training mix (our 799,306 generated rows and 200,694 public rows). The other fifteen languages each contain 41,934 rows (4.19\%).}
\label{fig:language-distribution}
\end{figure}

Seventeen languages are covered: English, Vietnamese, Chinese, Japanese, Lao, Tamil, Thai, Burmese, Filipino, Russian, Spanish, Malay, Indonesian, Portuguese, French, Korean and German. Seven of them (Vietnamese, Lao, Thai, Burmese, Filipino, Malay and Indonesian) are Southeast Asian languages, several of which are poorly represented in existing PII data; the others add widely used languages for script and family diversity (Latin, Cyrillic, Han, kana, Hangul, Thai, Lao, Myanmar and Tamil scripts). Figure~\ref{fig:language-distribution} shows the language distribution of the training mix, which also includes public rows (Section~\ref{sec:model}).

\subsection{Label ontology}
\label{sec:ontology}

We define nine labels informed by the identifiers in ISO/IEC 29100:2024 \cite{iso29100-2024}: \texttt{address} (street addresses, postal codes, coordinates, care locations); \texttt{company\_name} (hospitals, clinics, departments, companies); \texttt{date} (birth, admission, discharge and other dates tied to a person or encounter); \texttt{email\_address}; \texttt{human\_name}; \texttt{id\_number} (record, government, passport, licence, account numbers and IP addresses); \texttt{phone\_number} (including fax); \texttt{private\_url} (access-bearing links); and \texttt{secret} (passwords, keys, tokens, cookies, PINs, one-time passwords). The set covers the common textual identifier classes and adds credential categories that matter for portals and logs. It is a general PII ontology rather than a clinical de-identification standard; Table~\ref{tab:hipaa} relates it to the HIPAA Safe Harbor identifiers.

\begin{table}[!htbp]
\centering
\small
\caption{Safe Harbor identifier classes and the label of our ontology that covers them.}
\label{tab:hipaa}
\resizebox{\columnwidth}{!}{%
\begin{tabular}{p{0.52\columnwidth} p{0.38\columnwidth}}
\toprule
Safe Harbor identifier & Our label \\
\midrule
Names & human\_name \\
Geographic units smaller than a state & address \\
Dates (except year) tied to an individual & date (no shifting) \\
Telephone; fax numbers & phone\_number \\
Email addresses & email\_address \\
SSN; medical record; health-plan beneficiary; account; certificate/licence numbers & id\_number \\
Vehicle and device identifiers; serial numbers & id\_number (aliased, not specifically generated) \\
IP addresses & id\_number \\
Web URLs & private\_url (access-bearing only; public URLs deliberately untagged) \\
Biometric identifiers; full-face images & not covered / not text \\
Other unique identifying numbers or codes & id\_number (partial) \\
Ages over 89 & \textbf{not covered} (age is excluded by the prompt) \\
\bottomrule
\end{tabular}}
\end{table}

\subsection{Generation and verification pipeline}
\label{sec:pipeline}

\begin{figure*}[!t]
\centering
\includegraphics[width=1\textwidth]{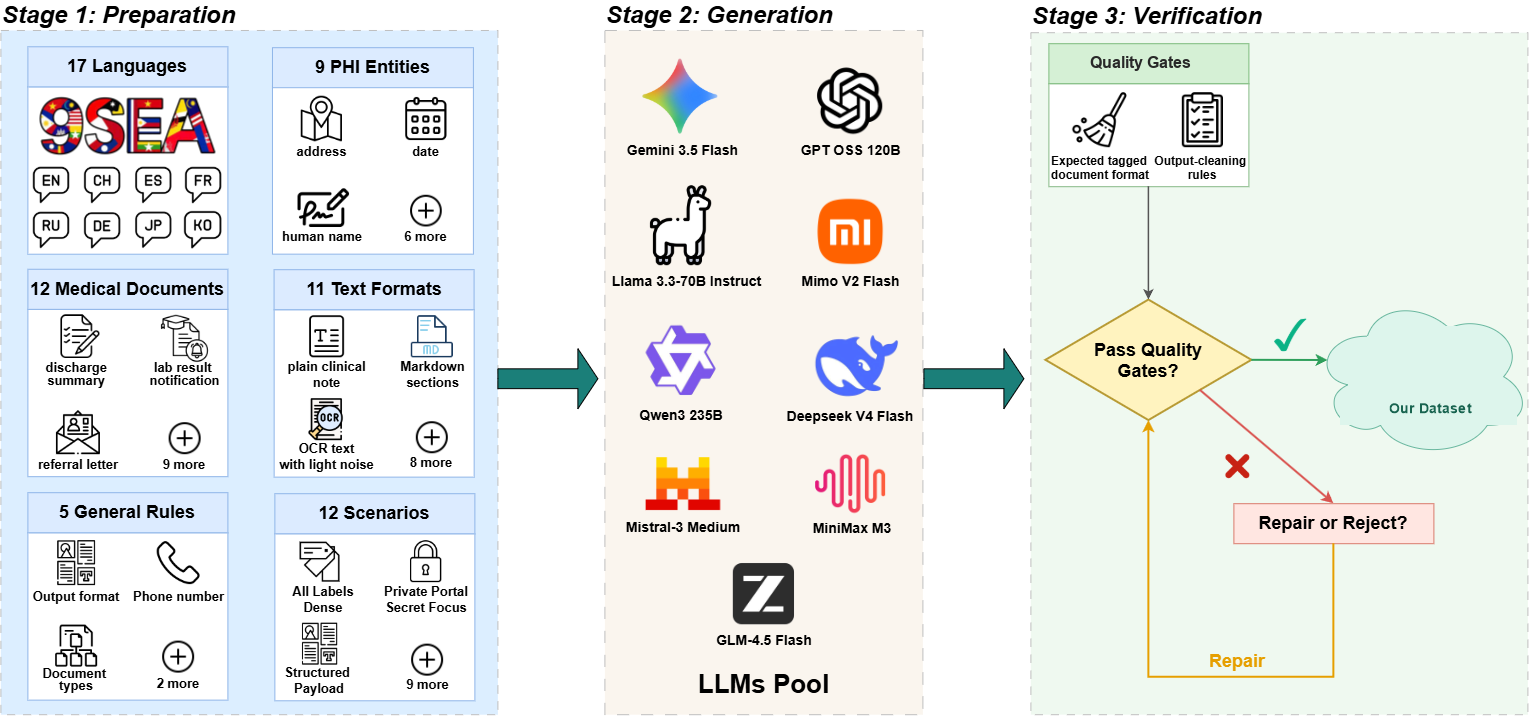}
\caption{Synthetic data generation and verification pipeline of the refined collection.}
\label{fig:data-generation-pipeline}
\end{figure*}

Figure~\ref{fig:data-generation-pipeline} summarizes the pipeline: Preparation, Generation, Verification. The user prompt is shown in Figure~\ref{fig:user-prompt} and the system prompt in Appendix~\ref{app:prompts}.

\paragraph{Stage 1: Preparation.} Each prompt combines six attribute groups: language; PII labels with required-label coverage; one of 12 medical document types; one of 11 text formats; general rules (output format, phone construction, private URLs, secrets, document/format compatibility); and one of 12 scenarios that set the operational context and label density (document types and text formats: Appendix~\ref{app:catalogues}). A span target of 3--20 is sampled per document. With probability 0.3 one of thirteen surface perturbations (spoken digits and email symbols, phonetic spelling, unusual spacing, line breaks inside values, truncation, partial masking, OCR confusions, Unicode confusables, concatenation, etc.; more details in Table~\ref{tab:perturbations}) is requested; adversarial scenarios request four or five distinct ones.

\begin{figure*}[!t]
\begin{tcolorbox}[enhanced, colback=MeddiesPromptBack, colframe=MeddiesPromptFrame, sharp corners, boxrule=0.9pt]
\begin{lstlisting}[style=meddiesprompt]
Generate one synthetic {language} {document_type} in {text_format} format.
Scenario: {scenario_name}
{scenario_description}
Hard constraints:
 1. Use at least {min_spans} tagged PII spans.
 2. Use at least {min_unique_labels} unique PII Labels labels.
 3. {required_instruction}   (e.g. "Required labels for this sample: <human_name>, <id_number>, ...")
 4. {private_secret_instruction}   (require / allow / do not force <private_url> and <secret>, depending on the scenario)
 5. {public_negative_rule}   (scenario- and language-specific hint about public, non-PII look-alikes)
 6. {adversarial_rule}   (scenario- and language-specific adversarial hint)
 7. {realism}   (default: keep the document realistic for healthcare/admin operations, but include no real people or real credentials)
 8. Output only the document content. Do not explain the labels.
 9. Before final output, self-check that every required label appears and every label uses `[value]<label>` exactly.
10. Do not output a checklist, legend, label inventory, or bare label such as `<human_name>` without a value.
11. If you use a phone number, bracket the whole number; do not tag short suffixes, extensions, or fragments as `<phone_number>`.
12. If you use `<private_url>`, bracket the full private URL including patient/account/session/token/auth query parameters.
13. If you use `<secret>`, bracket only the credential value itself: OTP, password, cookie, bearer token, API key, reset code, or session token.
14. Do not name or describe the document type or format in the output; produce the document content directly.
15. (only when edge cases were sampled) Apply ONE surface perturbation to at most 1-2 values, leaving the rest clean: {edge_case}
    -- or, for adversarial scenarios: Apply EACH of these {k} surface perturbations, each to a DIFFERENT value, and still leave 1-2 values completely clean for contrast: {edge_cases}
Required-label coverage before you answer: {label}=present, ...
Minimum coverage before you answer: at least {min_spans} bracketed tags and at least {min_unique_labels} unique labels.
Remember: label syntax must be `[value]<label>` and labels must be one of: <address>, <company_name>, <date>, <email_address>, <human_name>, <id_number>, <phone_number>, <private_url>, <secret>.
\end{lstlisting}
\end{tcolorbox}
\caption{User prompt template. Items 3--7 and 15 are filled from the sampled scenario, language profile and edge cases.}
\label{fig:user-prompt}
\end{figure*}

\paragraph{Stage 2: Generation.} Each prompt is sent to one of nine LLMs: Gemini 3.5 Flash \cite{googledeepmind2026gemini35flash}, GLM-4.5-Flash \cite{zeng2025glm}, Llama 3.3 70B Instruct \cite{meta2024llama33}, MiMo-V2-Flash \cite{xiao2026mimo}, Mistral Medium 3 \cite{mistralai2025mistralmedium3}, gpt-oss-120b \cite{agarwal2025gpt}, DeepSeek-V4-Flash \cite{deepseekai2026v4flash}, Qwen3-235B-A22B \cite{yang2025qwen3} and MiniMax M3 \cite{minimax2026m3}. Using several model families reduces dependence on one generator's style.

\paragraph{Stage 3: Verification and repair.} Each draft is parsed with a deterministic character-level parser and checked by 15 fatal and 2 non-fatal checks, plus two pre-validation rejections (Table~\ref{tab:gates}, Appendix~\ref{app:gates}): tag syntax, offsets that reproduce the span text, minimum length and span count, required labels, native-script presence, phone-number structure, and byte-identical duplicates. A conservative regular-expression repair step fixes explicit formatting defects (a bracket split, a missing bracket around a value, spacing before a tag) and the draft is revalidated before acceptance.

\subsection{Our Benchmark}

The benchmark contains 5,100 documents, 300 per language: 100 under standard conditions and 200 emphasizing adversarial surface forms, each combining four or five of the thirteen perturbations. It is produced by the same pipeline as the training rows (Section~\ref{sec:indomain}). Most of the benchmark was generated by GPT-5.4-mini, which is not one of the nine generators of the training data (Section~\ref{sec:pipeline}).

\subsection{Analysis}
\label{sec:audit}

We analyse the release with automated checks to measure its scale and label balance, its redundancy and overlap with the evaluation sets, and the defects it contains.

\paragraph{Scale and balance.} The refined collection alone has 317,846 rows and 5.3 million tagged spans (median 14--16 spans per document). All nine labels are well represented: from 395k spans (\texttt{address}) to 806k (\texttt{secret}), with 457k \texttt{company\_name}, 451k \texttt{email\_address} and 550k \texttt{phone\_number}. Median document length is about 1,100--1,300 characters for most languages and 680--730 for Chinese, Japanese and Korean.

Table~\ref{tab:audit} (Appendix~\ref{app:audit}) compares our rows with the public rows in the training mix: ours are longer and denser (16.7--20.8 tagged spans and a median of 1,144--2,180 characters per row, against 3.2--6.4 spans and 179--1,017 characters for the public corpora), and the same table reports the defects we found in each origin (Appendix~\ref{app:audit}).

\paragraph{Redundancy and overlap.} Near-duplication is rare: at estimated Jaccard $\geq0.8$ over character 5-grams, only 56 of 317,846 refined documents (0.02\%) have an earlier near-duplicate in their language (0.08\% after masking digits). None of the 18 evaluation sets (fifteen public, three in-house) contains a document that occurs in the training table, by exact hash or by whitespace- and case-normalized hash.

\section{Our Model}
\label{sec:model}

\begin{table*}[!t]
\centering
\caption{Exact-match entity-level micro-F1 (point estimates). Rows marked $\dagger$ have training rows of the same source and language in the training mix (Section~\ref{sec:composition}); the others have none in that language. All scores are recomputed from per-document predictions; 95\% intervals are in Appendix~\ref{app:ci}.}
\label{tab:benchmark-results}
{\resizebox{\textwidth}{!}{
\begin{tabular}{l*{5}{>{\centering\arraybackslash}m{2.45cm}}}
\toprule
\textbf{Benchmark} &
\textbf{Ours} &
\textbf{\shortstack{OpenAI Privacy\\Filter}} &
\textbf{GLiNER2} &
\textbf{\shortstack{SuperClinical\\Large 434M}} &
\textbf{\shortstack{LFM2.5 Encoder\\350M PII Detector}} \\
\midrule
\multicolumn{6}{l}{\textit{Public benchmarks}} \\
\multicolumn{6}{l}{\textit{(CredData: real source code; others synthetic)}} \\
CredData (en)$^\dagger$      & \textbf{0.590} & 0.584 & 0.357 & 0.149 & 0.263 \\
OpenPII (ko)  & \textbf{0.739} & 0.554 & 0.322 & 0.545 & 0.460 \\
OpenPII (ja)  & \textbf{0.750} & 0.549 & 0.366 & 0.493 & 0.479 \\
OpenPII (zh)  & \textbf{0.816} & 0.572 & 0.453 & 0.428 & 0.548 \\
OpenPII (de)  & \textbf{0.827} & 0.716 & 0.663 & 0.620 & 0.610 \\
OpenPII (ms)  & \textbf{0.839} & 0.640 & 0.692 & 0.547 & 0.587 \\
OpenPII (fr)  & \textbf{0.844} & 0.738 & 0.674 & 0.604 & 0.611 \\
OpenPII (fil) & \textbf{0.848} & 0.667 & 0.690 & 0.584 & 0.600 \\
OpenPII (id)  & \textbf{0.848} & 0.664 & 0.688 & 0.566 & 0.627 \\
OpenPII (pt)  & \textbf{0.855} & 0.756 & 0.692 & 0.687 & 0.651 \\
OpenPII (es)  & \textbf{0.870} & 0.794 & 0.732 & 0.683 & 0.655 \\
Gretel (en)$^\dagger$        & \textbf{0.872} & 0.697 & 0.588 & 0.715 & 0.729 \\
OpenPII (en)$^\dagger$  & \textbf{0.878} & 0.745 & 0.687 & 0.656 & 0.568 \\
OpenPII (vi)$^\dagger$  & \textbf{0.880} & 0.634 & 0.628 & 0.499 & 0.615 \\
Nemotron (en)$^\dagger$      & \textbf{0.956} & 0.764 & 0.694 & 0.894 & 0.573 \\
\textbf{Mean, all 15} & \textbf{0.827} & 0.672 & 0.595 & 0.578 & 0.572 \\
\textbf{\shortstack[l]{Mean, 10 OpenPII sets\\without same-source training rows}} &\textbf{0.824} & 0.665 & 0.597 & 0.576 & 0.583 \\
\midrule
\textbf{Our benchmark (in-domain)} & \textbf{0.878} & 0.540 & 0.476 & 0.244 & 0.269 \\
\bottomrule
\end{tabular}
}}
\end{table*}

\paragraph{Training data.} The model is trained on a one-million-row mix, counted from the source field of every row: our 799,306 generated rows plus 200,694 training-split rows of four public corpora for general-domain and code-log text (Nemotron-PII 99,885, Gretel PII 54,998, OpenPII 41,950, CredData 3,861). These public rows are not part of our dataset and are not redistributed. English has 255,503 rows (25.6\%) and Vietnamese 115,487 (11.5\%); each of the other fifteen languages has 41,934 (4.2\%), as shown in Figure~\ref{fig:language-distribution}.

We fine-tune a 350M-parameter encoder as a one-pass BIOES token classifier with a 37-class linear head (O plus four states for each of nine labels) and constrained Viterbi decoding \cite{lester-etal-2020-constrained}. One pass over the document makes inference much cheaper than autoregressive extraction, which in our earlier attempts (supervised fine-tuning followed by reinforcement learning) was slower and less accurate; we therefore adopted the token-classification formulation of OpenAI Privacy Filter \cite{openai2026privacyfilter}. The initialization, the full recipe and inference timing are in Appendix~\ref{app:training}; the run is a single seed.

\section{Evaluation}
\label{sec:eval}

\subsection{Settings}

We compare the model with four public systems: OpenAI Privacy Filter \cite{openai2026privacyfilter}, GLiNER2 \cite{zaratiana2025gliner2efficientmultitaskinformation}, SuperClinical-Large-434M-v1 \cite{openmed-pii-2026}, and LFM2.5-Encoder-350M-PII-Detector \cite{liquidai2026lfm25pii}. Native labels are folded into the nine-label ontology by the maps in Appendix~\ref{app:mappings}. We report exact-match entity-level micro-F1, pooled over documents: a prediction is a true positive only if label and both character boundaries match a gold span.
\begin{equation}
\begin{aligned}
P &= \frac{\mathrm{TP}}{\mathrm{TP}+\mathrm{FP}}, \qquad R = \frac{\mathrm{TP}}{\mathrm{TP}+\mathrm{FN}}, \\
F_1 &= \frac{2PR}{P+R} = \frac{2\mathrm{TP}}{2\mathrm{TP}+\mathrm{FP}+\mathrm{FN}}.
\end{aligned}
\end{equation}
On the human-annotated corpora we also report a \emph{relaxed} score (same label, at least one overlapping character, one-to-one matching), per-label scores, and 95\% percentile bootstrap intervals over documents (1,000 resamples). The label fold is authored by us and two baselines cannot emit \texttt{company\_name} at all (Table~\ref{tab:fold}).

\subsection{Evaluation on Public and In-Domain Benchmarks}

\paragraph{Public benchmarks.} Table~\ref{tab:benchmark-results} reports the main comparison. Our model obtains the highest score on all fifteen public benchmarks and is significantly ahead of the strongest baseline on fourteen of them, where the paired 95\% bootstrap intervals of the difference exclude zero (Appendix~\ref{app:ci}). Averaged over the fifteen sets, the model scores 0.827 against 0.672 for OpenAI Privacy Filter, a gain of 0.155. The advantage holds across scripts and language families, with the largest margins in Vietnamese (0.880 vs.\ 0.634), Chinese (0.816 vs.\ 0.572) and Japanese (0.751 vs.\ 0.549), and it persists on the ten OpenPII languages for which the training mix has no same-source rows (mean 0.824 vs.\ 0.665). Training on the mix also lifts its own initialization from 0.572 to 0.827 on average. For context, fourteen of the fifteen sets are LLM- or template-generated, so this is transfer across synthetic generators; five sets (marked $\dagger$) have same-source training rows in the mix, although no evaluation document appears in the training table (Section~\ref{sec:audit}); and CredData, built from credentials in real source code, is the hardest set for our model (0.590) and the only one where it is not significantly ahead.

\paragraph{In-domain benchmark.}\label{sec:indomain} On our benchmark the model scores 0.878 against 0.540 for the best baseline (mean of the standard and the adversarial split; pooled over the 5,100 documents 0.884 vs.~0.548). Because it shares its pipeline with the training data, we treat it as a regression test of the pipeline; its caveats and breakdowns are in Appendix~\ref{app:benchdeep}.

\subsection{Evaluation on Human-Annotated Corpora}
\label{sec:realtext}

We also evaluate, without adaptation, on two publicly shareable human-annotated corpora: GraSCCo\_PHI \cite{lohr2024deidentifying} (63 German clinical documents) and the MEDDOCAN test set \cite{marimon2019meddocan} (250 Spanish clinical case reports with surrogate PHI).

\paragraph{Definition differences account for part of the gap.} The scorer is exact, so any disagreement about what counts as PII, or how long a span is, is scored as an error, and the two guidelines define PHI differently from ours and from each other (Table~\ref{tab:defs}). To separate this from detection errors we re-score every system before and after two rules, applied identically to all systems (Table~\ref{tab:ladder}): (1) strip a leading honorific (``Frau'', ``Dr.'', ``Dra.'') from predicted names, because both guidelines annotate the name only; (2) ignore gold spans that our definition does not cover (countries, kinship words such as ``madre'', bare years and durations). Rule (1) alone raises the model from 0.583 to 0.683 on GraSCCo\_PHI, ahead of every baseline; both rules together give 0.701 and 0.770, ahead of the best baseline on both corpora (0.622 and 0.726). Hospital and institution names are within our definition and are not detected by the model or by any other system we tested, so we do not ignore them; doing so would give 0.715 and 0.791.

\paragraph{Results under native conventions.} Under each corpus's own annotation conventions the model reaches exact F1 of 0.583 and 0.720 (relaxed 0.787 and 0.817) and ranks second on both, behind SuperClinical-Large on GraSCCo\_PHI (0.619, overlapping intervals) and GLiNER2 on MEDDOCAN (0.745). Hospital and institution names, which our ontology covers, are detected by neither the model nor any other system tested, and the model was not fine-tuned on either guideline; the full tables are in Appendix~\ref{app:realtext}. In addition, these corpora are small and not diverse: one language and one genre each, raw running text only, with no structured or noisy formats. Nevertheless, these results still show performance transfer on these benchmarks, with large improvements over the model's initialization, LFM2.5-Encoder-350M-PII-Detector: +0.380 exact F1 on GraSCCo\_PHI (0.583 against 0.203) and +0.230 on MEDDOCAN (0.720 against 0.490).

\begin{table}[!htbp]
\centering
\small
\caption{Exact micro-F1 before and after applying the two rules (see text). Best per column in bold.}
\label{tab:ladder}
\resizebox{\columnwidth}{!}{%
\begin{tabular}{l cc cc}
\toprule
\multirow{2}{*}{System} & \multicolumn{2}{c}{GraSCCo\_PHI (de)} & \multicolumn{2}{c}{MEDDOCAN (es)} \\
 & Before & After & Before & After \\
\midrule
Ours & 0.583 & \textbf{0.701} & 0.720 & \textbf{0.770} \\
SuperClinical-Large & \textbf{0.619} & 0.622 & 0.589 & 0.553 \\
GLiNER2 & 0.419 & 0.487 & \textbf{0.745} & 0.726 \\
OpenAI Privacy Filter & 0.497 & 0.580 & 0.607 & 0.655 \\
LFM2.5-Encoder PII & 0.203 & 0.217 & 0.490 & 0.525 \\
\bottomrule
\end{tabular}}
\end{table}

\paragraph{Multilingual transfer.}\label{sec:multigrascco} We also test the model, without adaptation, on MultiGraSCCo \cite{baroud2026multigrascco}, the 63 GraSCCo documents translated into ten languages with their annotations preserved, scoring the direct-identifier layer with the same label map and exact-match scorer as above (Table~\ref{tab:multigrascco}). Ours is best in Russian, Polish, Ukrainian and Turkish (exact F1 0.638--0.713) and SuperClinical-Large is best in German, English and French, in some cases within our confidence intervals. The model also transfers across languages: relative to its initialization, LFM2.5-Encoder-350M-PII-Detector, it improves exact F1 in all ten languages, by 0.38 to 0.51 in the eight Latin- and Cyrillic-script languages, including Italian, Polish, Ukrainian and Turkish, which are absent from our training data, and by less in the two Arabic-script languages. All ten languages and all five systems are in Appendix~\ref{app:multigrascco}.

\begin{table}[!htbp]
\centering
\small
\caption{Exact micro-F1 on seven of the ten MultiGraSCCo languages (PHI layer): the languages in our training data ($^\ast$) and those where ours is best. Best per row in bold.}
\label{tab:multigrascco}
\resizebox{\columnwidth}{!}{%
\begin{tabular}{l ccccc}
\toprule
\raisebox{0.6em}{Language} & \raisebox{0.6em}{Ours} & \shortstack{OpenAI\\PF} & \raisebox{0.6em}{GLiNER2} & \shortstack{SuperClinical\\Large} & \shortstack{LFM2.5\\PII} \\
\midrule
German$^\ast$ & 0.583 & 0.497 & 0.419 & \textbf{0.616} & 0.205 \\
English$^\ast$ & 0.674 & 0.477 & 0.518 & \textbf{0.776} & 0.233 \\
French$^\ast$ & 0.715 & 0.601 & 0.516 & \textbf{0.717} & 0.237 \\
Russian$^\ast$ & \textbf{0.713} & 0.538 & 0.502 & 0.422 & 0.201 \\
Polish & \textbf{0.647} & 0.524 & 0.451 & 0.564 & 0.206 \\
Ukrainian & \textbf{0.701} & 0.488 & 0.464 & 0.439 & 0.200 \\
Turkish & \textbf{0.638} & 0.413 & 0.234 & 0.528 & 0.201 \\
\bottomrule
\end{tabular}}
\end{table}

\section{Conclusion}

We presented a large seventeen-language PII resource with a fully documented pipeline. A compact 350M-parameter model trained on it obtains the highest score on all fifteen public benchmarks, significantly ahead of the strongest baseline on fourteen of them, with a mean exact-match F1 of 0.827 against 0.672 for the strongest public baseline. Beyond the data and the model, we release the generation framework itself, which combines attribute-conditioned prompting with deterministic verification and repair and specifies languages, document types, formats and scenarios in catalogues, together with its prompts, checks and label maps. We hope these contributions will support reproducible multilingual PII research and the construction of further resources for new languages and domains.
\paragraph{Data and code availability.} Our dataset and benchmark comprise 799,306 generated rows; public-corpus rows used in training are not redistributed. The dataset, benchmark, model and code are planned for public release upon acceptance.

\section{Limitations}
\label{sec:limitations}

\textbf{Scope of the data.} All documents and labels in the resource are LLM-generated and have not been adjudicated by humans, so precision and recall against human annotation remain to be measured. Our automated audit (Appendix~\ref{app:audit}) shows that the data still contain some placeholder-like values, markup characters and untagged repeats, and that one scenario dominates the refined rows; these are the first targets of the next release.

\textbf{No real patient text.} Our results rest on synthetic and surrogate text; validation on real clinical records, for example i2b2/UTHealth under a data-use agreement, is future work.

\textbf{Ontology.} The nine labels are not a de-identification standard (Table~\ref{tab:hipaa}): they include no age label, do not separate dates to remove from dates to shift, and do not model quasi-identifiers such as a rare diagnosis. The model must not be used to certify that text is safe to share, and a deployed system should treat false negatives as the costly error.

\textbf{Evaluation.} Scores depend on our label fold, and our benchmark shares an annotator with the training data. Five public benchmarks have same-source rows in the training mix, and the model is a single training run without ablations, so gains cannot be attributed to one component. Multilingual transfer is uneven: results on Arabic and Persian are weak, and no Arabic-script language is in our training data (Section~\ref{sec:multigrascco}).

\textbf{Provenance.} Generator shares are uneven (Appendix~\ref{app:audit}), all generators were run with their default decoding settings, and no rejection rate is available for the first collection.

\textbf{Ethics and licences.} No real patient text and no personal data were used or released. The generated rows are released under CC BY-NC 4.0; the public corpora in the training mix keep their own licences and are not redistributed; the two human-annotated corpora are CC BY 4.0 and contain surrogate or alienated PHI only. The data and model are research resources.

\section*{Acknowledgments}
We thank Modal for providing compute credits.

\bibliography{references}

\clearpage
\appendix

\begin{table*}[!htbp]
\centering
\small
\caption{Map of the Appendix.}
\label{tab:appendix-map}
\resizebox{\textwidth}{!}{%
\begin{tabular}{l p{0.27\textwidth} p{0.30\textwidth} p{0.22\textwidth}}
\toprule
Section & Content & Supports & Cited from \\
\midrule
A. Generation Prompts & Complete system prompt (Figure~\ref{fig:system-prompt}; the user prompt, Figure~\ref{fig:user-prompt}, is in Sec.~3.3) & Reproducing Stage 1--2 of the pipeline & Sec.~3.3, Sec.~1 \\
B. Preparation Catalogues & 12 document types and 11 text formats (Tables~\ref{tab:doctypes}, \ref{tab:formats}), 12 scenarios with descriptions, required labels; general-domain catalogues & What can be generated vs.\ what was generated & Sec.~3.3, Sec.~3.5 \\
C. Surface Perturbations & The 13 requested perturbations (Table~\ref{tab:perturbations} in Sec.~3.3) & Adversarial half of the benchmark & Sec.~3.3, Sec.~3.4 \\
D. Verification Checks & All 15 fatal, 2 non-fatal and 2 pre-validation checks (Table~\ref{tab:gates} in Sec.~3.3) & Scope and limits of the quality gates & Sec.~3.3, Sec.~3.5 \\
E. Ontology Mappings & Label folds of the four baselines and of the two human-annotated corpora & Scoring protocol & Sec.~5.1, Sec.~5.3 \\
F. HIPAA Coverage & Safe Harbor identifiers vs.\ the nine labels (Table~\ref{tab:hipaa} in Sec.~3.2) & Scope of the ontology & Sec.~3.2, Sec.~7 \\
G. Audit Details & Per-generator, per-language and overlap tables (including the summary by origin, Table~\ref{tab:audit}) & Section 3.5 findings & Sec.~3.5 \\
H. Human-Annotated Corpora & Full results on both corpora (Tables~\ref{tab:realtext}, \ref{tab:realtext-labels}, \ref{tab:taxonomy}), definition comparison, alignment rules (the before/after table is in Sec.~5.3) & Section 5.3 claims & Sec.~5.3 \\
I. Training Recipe & Hyperparameters, data, seed and hardware of the single run & Ours (Section 4) & Sec.~4 \\
J. Benchmark Profile & Label counts, text formats, controlled and uncontrolled variables of our benchmark & Table~\ref{tab:corpora} (Sec.~3.4) & Sec.~3.4 \\
K. Public-benchmark intervals & 95\% bootstrap intervals of our model, the strongest baseline and their paired difference on the 15 public sets (Table~\ref{tab:ci}) & Table~\ref{tab:benchmark-results} claims & Sec.~5.2 \\
L. Benchmark breakdown & Our benchmark by text format, split and generator for all five systems (Tables~\ref{tab:bench-format}, \ref{tab:bench-gen}) & Sec.~5.5 claims & Sec.~5.5 \\
M. MultiGraSCCo & Results on all ten languages for ours and the four baselines (Tables~\ref{tab:multigrascco-exact}, \ref{tab:multigrascco-relaxed}) & Section~\ref{sec:multigrascco} claims & Sec.~\ref{sec:multigrascco} \\
\bottomrule
\end{tabular}}
\end{table*}

\section*{Contents}

{\small\setlength{\parindent}{0pt}\setlength{\parskip}{2pt}\raggedright
\hyperref[app:prompts]{\textbf{\ref{app:prompts}\enspace Generation Prompts}\nobreak\dotfill\nobreak \pageref{app:prompts}}\\
\hyperref[app:catalogues]{\textbf{\ref{app:catalogues}\enspace Preparation Catalogues}\nobreak\dotfill\nobreak \pageref{app:catalogues}}\\
\hspace*{1.2em}\hyperref[sub:doctypes]{\ref{sub:doctypes}\enspace Medical document types (12)\nobreak\dotfill\nobreak \pageref{sub:doctypes}}\\
\hspace*{1.2em}\hyperref[sub:formats]{\ref{sub:formats}\enspace Text formats (11)\nobreak\dotfill\nobreak \pageref{sub:formats}}\\
\hspace*{1.2em}\hyperref[sub:scenarios]{\ref{sub:scenarios}\enspace Scenarios (12)\nobreak\dotfill\nobreak \pageref{sub:scenarios}}\\
\hyperref[app:perturbations]{\textbf{\ref{app:perturbations}\enspace Surface Perturbations}\nobreak\dotfill\nobreak \pageref{app:perturbations}}\\
\hyperref[app:gates]{\textbf{\ref{app:gates}\enspace Verification Checks}\nobreak\dotfill\nobreak \pageref{app:gates}}\\
\hyperref[app:mappings]{\textbf{\ref{app:mappings}\enspace Ontology Mappings Used for Scoring}\nobreak\dotfill\nobreak \pageref{app:mappings}}\\
\hyperref[app:hipaa]{\textbf{\ref{app:hipaa}\enspace Coverage of HIPAA Safe Harbor Identifiers}\nobreak\dotfill\nobreak \pageref{app:hipaa}}\\
\hyperref[app:audit]{\textbf{\ref{app:audit}\enspace Audit Details}\nobreak\dotfill\nobreak \pageref{app:audit}}\\
\hspace*{1.2em}\hyperref[sub:defects]{\ref{sub:defects}\enspace Defect and diversity findings\nobreak\dotfill\nobreak \pageref{sub:defects}}\\
\hspace*{1.2em}\hyperref[sub:generators]{\ref{sub:generators}\enspace Generators, languages and evaluation overlap\nobreak\dotfill\nobreak \pageref{sub:generators}}\\
\hyperref[app:realtext]{\textbf{\ref{app:realtext}\enspace Human-Annotated Corpora: Results, Definitions and Alignment}\nobreak\dotfill\nobreak \pageref{app:realtext}}\\
\hspace*{1.2em}\hyperref[sub:schemes]{\ref{sub:schemes}\enspace How the three annotation schemes differ\nobreak\dotfill\nobreak \pageref{sub:schemes}}\\
\hspace*{1.2em}\hyperref[sub:results2]{\ref{sub:results2}\enspace Results on the two corpora\nobreak\dotfill\nobreak \pageref{sub:results2}}\\
\hspace*{1.2em}\hyperref[sub:align]{\ref{sub:align}\enspace Alignment rules\nobreak\dotfill\nobreak \pageref{sub:align}}\\
\hyperref[app:training]{\textbf{\ref{app:training}\enspace Training Recipe}\nobreak\dotfill\nobreak \pageref{app:training}}\\
\hyperref[app:benchmark]{\textbf{\ref{app:benchmark}\enspace Benchmark Profile}\nobreak\dotfill\nobreak \pageref{app:benchmark}}\\
\hyperref[app:multigrascco]{\textbf{\ref{app:multigrascco}\enspace MultiGraSCCo: All Ten Languages}\nobreak\dotfill\nobreak \pageref{app:multigrascco}}\\
\hyperref[app:benchdeep]{\textbf{\ref{app:benchdeep}\enspace Benchmark Breakdown}\nobreak\dotfill\nobreak \pageref{app:benchdeep}}\\
\hyperref[app:ci]{\textbf{\ref{app:ci}\enspace Public Benchmarks: Confidence Intervals}\nobreak\dotfill\nobreak \pageref{app:ci}}\\
}

\section{Generation Prompts}
\label{app:prompts}

Figure~\ref{fig:system-prompt} gives the complete system prompt used by the generator; the user prompt is Figure~\ref{fig:user-prompt} in Section~\ref{sec:pipeline}. Braced fields are filled from the sampled attributes. The system prompt is shared by all generators; only the language, domain hint and public-URL hint vary. The prompt text is copied from the released generation code.

\begin{figure*}[!htbp]
\begin{tcolorbox}[enhanced, colback=MeddiesPromptBack, colframe=MeddiesPromptFrame, sharp corners, boxrule=0.9pt]
\begin{lstlisting}[style=meddiesprompt]
You generate fully synthetic PII training documents for a PII Labels BIOES span detector.
Output only the document. No preface, no explanation, no markdown code fence unless the requested document itself is structured text.
Language: {language}.  Domain: {domain}.
Allowed positive labels, and only these labels: <address> <company_name> <date> <email_address> <human_name> <id_number> <phone_number> <private_url> <secret>
Annotation format is exactly `[specific synthetic value]<label>`.
Every positive label marker must be immediately preceded by a square-bracketed value. Never write `value<label>` or dangling `<label>` markers.
For JSON, XML, HL7, tables, or logs, the string field value itself must contain `[value]<label>`.
Good examples: `[Maya Green]<human_name>`, `[maya.green@example.com]<email_address>`,
  `[https://portal.example.test/patients/P-104/results?token=tok_8f91]<private_url>`, `[otp-482-119]<secret>`.
Bad examples: `Maya Green<human_name>` (value not bracketed); `<human_name>Maya Green` (label before value);
  `[Maya Green] <human_name>` (space splits value and label); `[Maya Green]<human_name` (missing `>`).
Human names are always PII. Dates are PII only when tied to a patient/person/encounter/credential; public guideline/article/copyright dates are not PII.
Private URLs are patient portals, signed links, private records/results/claims/messages, or URLs with patient/session/token/auth parameters. {public_url_rule}
Secrets are passwords, OTPs, PINs, API keys, bearer/access/auth/session/CSRF tokens, cookies, reset links/codes.
Phone numbers must be tagged as the full phone number, including country/area code and extension when present. Do not tag a 4-digit extension or suffix as a standalone `<phone_number>`.
Tag the MINIMAL entity value only - never a whole sentence, title, heading, descriptive clause, or document label. `company_name` is the institution name itself, not a sentence describing it.
When digits are spoken aloud or OCR-transcribed as words, still tag by WHAT the number is: a contact phone number is `<phone_number>`; a patient/record/claim/insurance/account identifier is `<id_number>`. Tag the full spoken span but do not absorb trailing narration words.
Never tag vitals, lab measurements, dosages, medical terms, diagnoses, procedure names, medication names, age, sex/gender, or public/non-person facts.
All names, addresses, IDs, URLs, emails, tokens, dates, and institutions must be fictional; use reserved example domains (example.com, example.org, example.net, example.test) for emails and URLs.
Make every phone number and identifier look realistic and varied; never use lazy sequential or repeated digit runs such as 0987654321, 0123456789, or 1111111111.
\end{lstlisting}
\end{tcolorbox}
\caption{System prompt of the generator.}
\label{fig:system-prompt}
\end{figure*}

\section{Preparation Catalogues}
\label{app:catalogues}

The generator draws a document type, a text format and a scenario for every sample. Only the \emph{names} of document types and formats appear in the prompt (``Generate one synthetic \{language\} \{document type\} in \{text format\} format''); the released code gives them no per-item definition, required structure or compatibility rule. The ``gloss'' columns below are therefore our \emph{descriptive} summaries for the reader, not specifications that the generator or the validator used. Scenarios, in contrast, carry a description, required labels and minimum counts in the catalogue, reproduced here. ``Rows'' is the number of rows of the refined collection (317,846) recorded with that value.

\subsection{Medical document types (12)}
\label{sub:doctypes}

\begin{table*}[!htbp]
\centering
\small
\caption{Medical document types. Gloss: descriptive summary by the authors (not a generator-side definition).}
\label{tab:doctypes}
\resizebox{\textwidth}{!}{%
\begin{tabular}{r l p{0.55\textwidth} r}
\toprule
\# & Document type & Gloss & Rows \\
\midrule
1 & Discharge summary & End-of-stay summary of diagnosis, treatment and follow-up, with patient, encounter and clinician identifiers. & 30,238 \\
2 & Outpatient note & Clinician note from a non-admitted visit (reason, findings, plan) with patient and visit details. & 19,492 \\
3 & Lab-result notification & Message telling a patient or clinician that results are available, usually with a result link and an accession or patient ID. & 29,774 \\
4 & Patient portal access log & System record of logins, result views and downloads with user, timestamp and session data. & 25,540 \\
5 & Insurance claim note & Payer or provider note about a claim, with policy, member and claim identifiers and contact data. & 24,464 \\
6 & Referral letter & Letter from one clinician to another clinician or service giving patient identity and reason for referral. & 23,634 \\
7 & Pharmacy refill request & Request or confirmation of a prescription refill with patient, prescriber, pharmacy and contact data. & 33,222 \\
8 & Appointment reminder & Scheduling message with patient name, date, location and contact numbers. & 15,951 \\
9 & Telehealth triage note & Remote-consultation note recording symptoms and disposition, with patient contact and session identifiers. & 26,636 \\
10 & Medical bill / invoice & Statement of charges with patient, account and payer details. & 19,328 \\
11 & FHIR-like Patient + Encounter resource & Structured JSON shaped like HL7 FHIR Patient and Encounter resources (identifier, name, telecom, address, period). & 10,981 \\
12 & HL7-style admission / update message & Pipe-delimited message in the style of HL7 v2 ADT events (MSH, PID, PV1 segments). & 21,297 \\
\bottomrule
\end{tabular}}
\end{table*}

\subsection{Text formats (11)}
\label{sub:formats}

\begin{table*}[!htbp]
\centering
\small
\caption{Text formats. Gloss: descriptive summary by the authors (not a generator-side definition). Bucket: canonical surface form used for corpus statistics.}
\label{tab:formats}
\resizebox{\textwidth}{!}{%
\begin{tabular}{r l l p{0.42\textwidth} r}
\toprule
\# & Text format & Bucket & Gloss & Rows \\
\midrule
1 & Plain clinical note & PLAIN\_TEXT & Running prose, lightly sectioned. & 13,423 \\
2 & Markdown sections & MARKDOWN & Headings, lists and tables in Markdown. & 29,316 \\
3 & Messy nurse / administrative note & MESSY\_NOTE & Informal note with abbreviations, fragments and inconsistent punctuation. & 36,000 \\
4 & Table-like rows & TABULAR & Rows with delimiters or aligned columns. & 34,203 \\
5 & JSON-like object & JSON & JSON-style key--value object; not guaranteed to be valid JSON. & 42,355 \\
6 & FHIR-like JSON & FHIR\_JSON & JSON shaped like a FHIR resource. & 25,827 \\
7 & HL7-like pipe-delimited text & HL7\_V2 & Segments and fields separated by pipes. & 26,583 \\
8 & Patient portal audit log & PLAIN\_TEXT & One event per line (time, actor, action). & 25,782 \\
9 & Voice-scribe transcript & PLAIN\_TEXT & Dictated or ambient-scribe transcript, with spoken numbers. & 28,734 \\
10 & Lightly degraded OCR text & MESSY\_NOTE & Text with scanner-style character errors. & 25,461 \\
11 & Clinical dialogue transcript & PLAIN\_TEXT & Speaker-turn conversation between clinician and patient. & 10,593 \\
\bottomrule
\end{tabular}}
\end{table*}

Formats describe surface form only; the document identity (note, log, letter) is carried by the document type. The evaluator's corpus statistics fold each prose name onto a canonical bucket (column ``Bucket'').

\paragraph{General-domain catalogues.} The released code also defines two non-medical profiles, which explain why the metadata of the refined collection contain 45 document-type and 25 format strings. General document types: customer support ticket; billing dispute; shipping update; school enrollment note; telecom account note; insurance claim message; HR onboarding form; account recovery email. Code/log document types: application audit log; \texttt{.env}-style configuration snippet; support ticket with request headers; account recovery trace; JSON security event; API error report. General formats: plain support note; email thread; Markdown sections; table-like rows; form fields; chat transcript. Code/log formats: JSON-like object; \texttt{.env} key--value block; HTTP headers; application log lines; YAML-like config; support ticket with code block.

\subsection{Scenarios (12)}
\label{sub:scenarios}

A scenario fixes the operational context, the labels that must occur and the minimum label and span counts; two scenarios are adversarial (several perturbations per document). Descriptions are condensed from the catalogue text that is inserted into the user prompt. In the refined collection, 3.6\% of rows record no scenario, and two further values occur that are not among the twelve (\texttt{local\_new\_data\_clinical\_forms}, 0.09\%; \texttt{voice\_ocr\_transcription}, $<$0.01\%).

\begin{table*}[!htbp]
\centering
\small
\caption{Scenario catalogue.}
\label{tab:scenarios}
\resizebox{\textwidth}{!}{%
\begin{tabular}{r p{0.17\textwidth} p{0.34\textwidth} p{0.17\textwidth} r }
\toprule
\# & Scenario & Description & Required labels & Min.\ labels/spans \\
\midrule
1 & All Labels Dense & Dense administrative-medical document containing every label at least once, including a private portal URL and a secret. & all nine & 9 / 12 \\
2 & Private Portal Secret Focus & Portal or lab-result workflow emphasizing signed links, reset tokens, OTPs, cookies or API keys, plus ordinary PII. & private\_url, secret, human\_name, id\_number & 6 / 10 \\
3 & Adversarial Obfuscation & Voice-scribe, OCR, chat and messy admin text with several values in hard-to-parse, language-native surface forms and 1--2 values left clean. & private\_url, secret, email\_address, phone\_number & 6 / 10 \\
4 & Public Negative Contrast & Patient PII tagged beside untagged public guideline dates, copyright dates and public hospital or article URLs. & private\_url, secret, date, human\_name & 6 / 10 \\
5 & Structured Payload & JSON, FHIR-like JSON, HL7-like text, XML, tables or portal logs; labels only inside string values, clinical facts untagged. & private\_url, secret, id\_number, company\_name & 6 / 10 \\
6 & General Consumer Admin Support & Account support, billing, shipping, insurance, school, telecom or e-commerce document. & human\_name, email\_address, id\_number & 5 / 5 \\
7 & HR, Employer, School, Insurance or Telecom Admin & HR, employer, school, insurance, telecom, banking-adjacent or civic administration with IDs, names, addresses and organizations. & human\_name, id\_number, email\_address & 5 / 6 \\
8 & Code, Log, Security or Account Recovery & Code, log, ticket, account-recovery or incident text using credentials, headers, cookies, API keys or tokens. & secret, id\_number & 4 / 5 \\
9 & Adversarial Formatting & General (non-medical) adversarial PII formatting; messy values with some left clean. & email\_address, phone\_number, id\_number & 5 / 6 \\
10 & Structured General Payload & General JSON, YAML, key--value logs, CSV-like rows, headers or audit records; string values only. & id\_number, email\_address & 4 / 5 \\
11 & One-Hop Deferred Clue & Surrounding text explains why a link, token, ID or date belongs to a named person or account. & human\_name, id\_number & 5 / 6 \\
12 & Public Negative URL and Data Contrast & General-domain PII beside untagged public URLs, documentation links, status pages and public dates. & human\_name, date, email\_address & 5 / 6 \\
\bottomrule
\end{tabular}}
\end{table*}

\paragraph{Sampling.} For each sample the framework draws a language, document type, format and scenario, and a span target. With probability 0.3 one perturbation from Table~\ref{tab:perturbations} is requested; adversarial scenarios request four or five distinct perturbations, each applied to a different value. Four of the twelve scenarios, including both general-domain adversarial and structured scenarios, do not occur at all in the refined collection, and one scenario accounts for 92.5\% of its rows.

\section{Surface Perturbations}
\label{app:perturbations}

The thirteen surface perturbations that the prompt can request for individual values are listed in Table~\ref{tab:perturbations} (Section~\ref{sec:pipeline}). They are the only adversarial mechanism of the pipeline and define the adversarial half of our benchmark.

\begin{table*}[!htbp]
\centering
\small
\caption{The thirteen surface perturbations requested by the prompt (paraphrased from the released catalogue). The same set defines the adversarial half of our benchmark.}
\label{tab:perturbations}
\resizebox{\textwidth}{!}{%
\begin{tabular}{r p{0.88\textwidth}}
\toprule
\# & Perturbation \\
\midrule
1 & Spoken email symbols: write \texttt{@} and \texttt{.} as the language's spoken words, tagging the whole spoken form. \\
2 & Number words: spell phone or ID digits as number-words, mixing numerals and words; tag by what the number is. \\
3 & Phonetic spelling: spell an identifier with a NATO or native spelling alphabet. \\
4 & Entity-internal line break: split a phone, email, URL or ID across a line break inside the tagged value. \\
5 & Odd internal spacing inside one value (\texttt{04 12 3 4 56 78}, \texttt{j smith @ example . com}). \\
6 & Concatenated values: run one value into an adjacent value or word with no separator; each value keeps its own exact span. \\
7 & Truncation: clip one value mid-string as a column width or log limit would. \\
8 & Full-width or look-alike punctuation in place of separators. \\
9 & Unicode confusables: swap one or two Latin letters for visually identical Cyrillic or Greek letters. \\
10 & OCR character confusion ($l\leftrightarrow1$, $O\leftrightarrow0$, $5\leftrightarrow S$, $rn\leftrightarrow m$, $B\leftrightarrow8$). \\
11 & Emoji in place of a separator or label word near a value; the emoji is not tagged. \\
12 & Minimal name fragment: given name only, initials, or a one-token surname. \\
13 & Partial masking: only some characters survive (\texttt{098****4567}); the whole masked value is tagged. \\
\bottomrule
\end{tabular}}
\end{table*}

\section{Verification Checks}
\label{app:gates}

Table~\ref{tab:gates} (Section~\ref{sec:pipeline}) lists every check applied by the released validator and acceptance loop. The validator returns a list of error codes; a draft is rejected if any \emph{fatal} code is present. Two phone-related codes are non-fatal. Two further rejections occur before validation. The \emph{repair} step (Section~\ref{sec:pipeline}) runs before revalidation and records \texttt{had\_label\_repairs}.

None of these checks inspects whether a tagged value is realistic, whether every identifier in the document is tagged, whether the same value is tagged consistently within a document, or whether the document is a faithful instance of its declared format (Section~\ref{sec:audit}).

\begin{table*}[!htbp]
\centering
\small
\caption{Checks in the released validator and acceptance loop. ``Fatal'' codes reject a draft.}
\label{tab:gates}
\resizebox{\textwidth}{!}{%
\begin{tabular}{r l c p{0.60\textwidth}}
\toprule
\# & Code & Fatal & Condition (what it does and does not establish) \\
\midrule
1 & \texttt{empty\_output} & yes & Nothing remains after stripping reasoning blocks, code fences and wrappers. \\
2 & \texttt{assistant\_meta\_text} & yes & The first 160 characters match an assistant-preface pattern. \\
3 & \texttt{no\_inline\_tags} & yes & The string \texttt{]<} does not occur. \\
4 & \texttt{leftover\_or\_unknown\_tags} & yes & A bracketed fragment followed by a tag survives parsing (unknown label name). \\
5 & \texttt{leftover\_label\_marker} & yes & A bare \texttt{<label>} marker remains in the output text. \\
6 & \texttt{malformed\_leftover\_label\_marker} & yes & A malformed marker such as \texttt{<label} remains. \\
7 & \texttt{too\_few\_spans} & yes & Fewer than two parsed spans (the prompt requests 3--20). \\
8 & \texttt{missing\_required\_labels} & yes & A label required by the sampled scenario never occurs. \\
9 & \texttt{too\_short} & yes & Fewer than 240 characters (120 for Chinese, Japanese, Korean). \\
10 & \texttt{too\_long} & yes & More than 16,000 characters. \\
11 & \texttt{bad\_span\_offsets} & yes & The recorded offsets do not reproduce the span text. \\
12 & \texttt{empty\_span} & yes & A span is empty or whitespace. \\
13 & \texttt{templated\_phone\_span} & yes & A phone span with $\ge$6 digits that is one repeated digit, or an ascending or descending run over its \emph{entire} length. It does not reject partial runs (\texttt{090-1234-5678}), reserved exchanges (\texttt{555}), or repeated values across documents. \\
14 & \texttt{missing\_native\_script} & yes & No character from the language's script (Vietnamese: diacritics) occurs; applies to vi, th, lo, my, ta, zh, ja, ko, ru. \\
15 & \texttt{duplicate\_text} & yes & The SHA-256 of the exact text equals that of an accepted document. Detects byte-identical documents only. \\
\midrule
16 & \texttt{short\_phone\_span} & no & A phone span has 1--6 digits. \\
17 & \texttt{underbounded\_phone\_span} & no & The surrounding phone-like window has more digits than the tagged span. \\
\midrule
-- & \texttt{api\_error} & pre & The provider call failed or returned no content. \\
-- & unknown scenario & pre & The scenario metadata needed for validation is unavailable. \\
\bottomrule
\end{tabular}}
\end{table*}

\section{Ontology Mappings Used for Scoring}
\label{app:mappings}

All systems are scored in the nine-label ontology. Each system's native labels are folded by the evaluator's maps (Table~\ref{tab:fold}). A system is credited only for labels its fold can produce; the last column lists nine-label classes it cannot emit under this fold. Because the fold is authored by us, every number depends on it; the machine-readable maps are released with the evaluation code.

\begin{table*}[!htbp]
\centering
\small
\caption{Native-label folds applied to each baseline.}
\label{tab:fold}
\resizebox{\textwidth}{!}{%
\begin{tabular}{l >{\raggedright\arraybackslash}p{0.52\textwidth} >{\raggedright\arraybackslash}p{0.2\textwidth}}
\toprule
System & Native labels $\rightarrow$ our label & Cannot emit \\
\midrule
OpenAI Privacy Filter & \texttt{private\_person}$\rightarrow$human\_name; \texttt{private\_address}$\rightarrow$address; \texttt{private\_date}$\rightarrow$date; \texttt{private\_email}$\rightarrow$email\_address; \texttt{private\_phone}$\rightarrow$phone\_number; \texttt{account\_number}$\rightarrow$id\_number; \texttt{private\_url}$\rightarrow$private\_url; \texttt{secret}$\rightarrow$secret & company\_name \\
GLiNER2 (requested schema) & person, full/first/last name$\rightarrow$human\_name; address, street, city, state, postal code, country$\rightarrow$address; sensitive/document/expiration/transaction date$\rightarrow$date; email$\rightarrow$email\_address; phone$\rightarrow$phone\_number; government/national/passport/driver's-licence/tax ID, account number$\rightarrow$id\_number; password, API key, access token, recovery code$\rightarrow$secret & company\_name, private\_url \\
LFM2.5-Encoder-350M-PII-Detector & 40 native types, e.g. person name, username$\rightarrow$human\_name; SSN, national ID, passport, driver's licence, tax ID, credit card, IBAN, bank account, IP, MAC, device ID, medical record, health-plan ID, case number$\rightarrow$id\_number; DOB$\rightarrow$date; email; phone; address, postal code, GPS$\rightarrow$address; organisation$\rightarrow$company\_name; API key, password, private key, JWT, connection string, login credentials$\rightarrow$secret; URL$\rightarrow$private\_url; amount, condition, medication, religion, political, orientation, health status $\rightarrow$ dropped & --- \\
SuperClinical-Large-434M & generic alias table of the evaluator (108 aliases: e.g. first/last/full name, username$\rightarrow$human\_name; street, city, state, country, postcode, coordinate$\rightarrow$address; hospital, organisation$\rightarrow$company\_name; DOB, date/time$\rightarrow$date; \ldots); \texttt{url}$\rightarrow$private\_url; \texttt{cvv}$\rightarrow$secret & --- (depends on emitted native types) \\
\bottomrule
\end{tabular}}
\end{table*}

\paragraph{Human-annotated corpora.} Source labels of GraSCCo\_PHI and MEDDOCAN are mapped as in Table~\ref{tab:realmap}. Gold spans whose source label has no counterpart (age, profession, sex, honorific title, \ldots) are \emph{ignored regions}: they are not scored and a prediction that overlaps only such a region is discarded rather than counted as a false positive.

\begin{table*}[!htbp]
\centering
\small
\caption{Source-label mapping for the two human-annotated corpora.}
\label{tab:realmap}
\resizebox{\textwidth}{!}{%
\begin{tabular}{l >{\raggedright\arraybackslash}p{0.40\textwidth} >{\raggedright\arraybackslash}p{0.46\textwidth}}
\toprule
Our label & GraSCCo\_PHI (German) & MEDDOCAN (Spanish) \\
\midrule
human\_name & NAME\_PATIENT, NAME\_DOCTOR, NAME\_RELATIVE, NAME\_EXT & NOMBRE\_SUJETO\_ASISTENCIA, NOMBRE\_PERSONAL\_SANITARIO, FAMILIARES\_SUJETO\_ASISTENCIA \\
date & DATE & FECHAS \\
address & LOCATION\_STREET, \_ZIP, \_CITY, \_COUNTRY & CALLE, TERRITORIO, PAIS \\
company\_name & LOCATION\_HOSPITAL, LOCATION\_ORGANIZATION & HOSPITAL, CENTRO\_SALUD, INSTITUCION \\
id\_number & ID & ID\_ASEGURAMIENTO, ID\_CONTACTO\_ASISTENCIAL, ID\_SUJETO\_ASISTENCIA, ID\_TITULACION\_PERSONAL\_SANITARIO, ID\_EMPLEO\_PERSONAL\_SANITARIO \\
phone\_number & CONTACT\_PHONE, CONTACT\_FAX & NUMERO\_TELEFONO, NUMERO\_FAX \\
email\_address & CONTACT\_EMAIL & CORREO\_ELECTRONICO \\
private\_url, secret & (none annotated) & (none annotated) \\
\emph{ignored} & NAME\_TITLE, AGE, PROFESSION, NAME\_USERNAME & EDAD\_SUJETO\_ASISTENCIA, SEXO\_SUJETO\_ASISTENCIA, PROFESION, OTROS\_SUJETO\_ASISTENCIA \\
\bottomrule
\end{tabular}}
\end{table*}

\section{Coverage of HIPAA Safe Harbor Identifiers}
\label{app:hipaa}

Table~\ref{tab:hipaa} (Section~\ref{sec:ontology}) relates the nine labels to the eighteen identifier classes of the HIPAA Safe Harbor method (45 CFR 164.514(b)(2)). Two labels, \texttt{company\_name} and \texttt{secret}, go beyond Safe Harbor. The ontology was designed from ISO/IEC 29100 identifiers for general PII extraction rather than from Safe Harbor, so coverage is partial and should not be read as HIPAA compliance.

\section{Audit Details}
\label{app:audit}

\subsection{Defect and diversity findings}
\label{sub:defects}

The verification checks of Section~\ref{sec:pipeline} establish that a sample has the declared \emph{shape}; they do not establish that values are realistic, that tagging is complete or consistent, or that the data are diverse. We therefore audit the release with automated heuristics. No human adjudication was performed, so all figures are estimates of specific, mechanically detectable defects and should be read as lower bounds on total annotation error. The scripts are released.

\paragraph{Summary of known defects.} The structural checks of Section~\ref{sec:pipeline} were not designed to judge whether values are realistic or tagging is complete. In the refined collection, a share of values is placeholder-like (e.g.\ 58.4\% of English phone numbers use 555 as area code or exchange, and sequential digit fillers occur in several languages), 4.0\% of tagged spans contain markup characters, and 40.3\% of documents leave a tagged value untagged elsewhere in the same document; one scenario accounts for 92.5\% of the refined rows. We report them by origin in Table~\ref{tab:audit} and in full below and plan to address them in the next release (locale-sampled values, a within-document consistency check, scenario rebalancing, human adjudication); Section~\ref{sec:limitations} discusses their consequences for use.

\paragraph{Protocol.} We audit (i) all 317,846 rows of the released refined collection, with per-generator and per-language breakdowns, and (ii) the full training mix by origin. Heuristics: a span is \emph{markup-contaminated} if its text contains \texttt{[}, \texttt{]} or a newline; \emph{over-long} if it exceeds 80 characters and is not an address, URL, secret or organization; a phone is \emph{placeholder-like} if it has 555 as area code or exchange or contains a sequential run such as 1234567; a name is a \emph{canonical placeholder} if it is on a short list of well-known specimen names; a document has a \emph{bare repeat} if a value of $\geq4$ characters tagged as name, ID, phone or email also occurs elsewhere in the same document without a tag. Near-duplicates use MinHash-LSH over character 5-grams (64 permutations, estimated Jaccard $\geq0.8$, within language).

\paragraph{Findings.} Table~\ref{tab:audit} summarizes the full table by origin; the rest of this appendix gives per-generator, per-language and overlap detail.

\begin{table*}[!htbp]
\centering
\small
\caption{Automated audit of the training mix by origin. Markup: spans containing \texttt{[}, \texttt{]} or a newline. 555: phones with 555 as area code or exchange. Seq.: phones containing a sequential digit run. Placeholder: canonical specimen names per 1,000 name spans. Bare repeat: documents in which a tagged name, ID, phone or email also occurs untagged. Percentages are of spans, phone spans or documents as appropriate. Origins are ordered from the most to the fewest defects, by the mean rank over the six defect columns (Markup spans to Bare repeat; a dash is skipped).}
\label{tab:audit}
\resizebox{\textwidth}{!}{%
\begin{tabular}{l r r r r r r r r r}
\toprule
 & & Spans & Median & Markup & Docs with & & & Placeholder & Bare \\
Origin & Rows & /row & chars & spans (\%) & markup (\%) & 555 (\%) & Seq. (\%) & names (\textperthousand) & repeat (\%) \\
\midrule
Ours, refined & 226,546 & 16.7 & 1,144 & 4.06 & 25.2 & 8.0 & 16.6 & 48.6 & 40.4 \\
Ours, initial & 572,760 & 20.8 & 2,180 & 0.02 & 0.2 & 8.0 & 31.3 & 20.6 & 27.9 \\
CredData & 3,861 & 3.2 & 1,017 & 9.55 & 11.9 & -- & -- & 0.0 & 0.0 \\
Nemotron-PII & 99,885 & 6.4 & 774 & 0.04 & 0.1 & 2.8 & 3.2 & 13.2 & 4.9 \\
Gretel PII & 54,998 & 4.2 & 179 & 0.00 & 0.0 & 0.1 & 0.0 & 0.0 & 0.0 \\
OpenPII & 41,950 & 5.7 & 277 & 0.00 & 0.0 & 0.0 & 0.0 & 0.0 & 0.1 \\
\bottomrule
\end{tabular}}
\end{table*}

\textbf{The values are often locale placeholders, and the phone check does not catch them.} In the refined collection, 58.4\% of English phone numbers use 555 as area code or exchange (25\% fall in the 555-01XX range reserved for fiction), although the validator's phone check is described as rejecting templated numbers: it rejects only a digit run over the \emph{entire} number (Appendix~\ref{app:gates}). Sequential filler such as \texttt{090-1234-5678} appears in 38.0\% of Korean, 36.2\% of Indonesian, 30.7\% of Japanese and 26.5\% of Vietnamese phone spans, and well-known specimen names (for example the Japanese pair 山田太郎 and 佐藤花子) account for 16.1\% of Chinese, 11.7\% of Vietnamese and 10.6\% of Japanese name spans. Entity diversity is low: the five most frequent names cover 46.3\% of Korean, 36.8\% of Chinese and 31.4\% of Indonesian name spans, and the unique-value ratio for names is 0.09 in Korean and 0.12 in Japanese. The initial collection is worse on phones (31.3\% sequential filler) and has $>$68k canonical placeholder names.

\textbf{Markup contaminates a quarter of the refined documents.} In the refined collection 4.0\% of spans (25.3\% of documents) contain a bracket or newline from the markup, for example a name span that begins at ``Patient: Yes. My name is [Maya Green''. The initial collection has 0.02\%, so the defect is specific to the refined collection. It is concentrated in some generators (qwen3-32b 8.0\% of spans, gpt-oss-120b 5.6\%, versus 1.0\% for gpt-5.4-mini; Table~\ref{tab:generators}), but none of the checks in Appendix~\ref{app:gates} looks inside a span for markup characters, and the repair rate itself (34.6\% of rows, 71.9\% for qwen3-32b) shows how often the first draft failed.

\textbf{Tagging is inconsistent within documents.} In 40.3\% of the refined documents and 27.9\% of the initial ones, a tagged name, ID, phone or email also occurs elsewhere in the same document without a tag. Some of these may be mentions that the guidelines legitimately leave bare, but otherwise the same string is both positive and negative supervision within one example. We have not adjudicated them.

\textbf{Diversity is narrower than the catalogue suggests.} In the refined collection 92.5\% of rows come from one of the twelve scenarios (\texttt{private\_portal\_secret\_focus}), 3.6\% have no scenario recorded, and the other eleven share 3.9\%. The metadata contain 45 document-type and 25 text-format strings against 12 and 11 declared, because general and code-log profiles and legacy values are mixed in. Near-duplication, in contrast, is rare: at estimated Jaccard $\geq0.8$ over character 5-grams only 56 of 317,846 refined documents (0.02\%) have an earlier near-duplicate within their language (0.08\% after masking digits), so exact hashing is weak but not the main problem.

\textbf{Encoding damage.} 824 refined documents contain the replacement character U+FFFD, 736 of them in Lao and Burmese.

\textbf{No training/evaluation text overlap, but shared entity values.} None of the 18 evaluation sets (fifteen external, three in-house) has a document that occurs in the training table, by exact hash or by whitespace- and case-normalized hash. However, the \emph{values} are shared: the share of gold spans whose exact (label, value) pair also occurs among training spans is 41.3\% for Nemotron against Nemotron training rows, 50.7\% for Vietnamese OpenPII against its Vietnamese training rows, 31.7\% for English OpenPII, and 30.8\% for Nemotron against self-generated rows; for CredData it is 0.0\%. Across the fifteen external sets the rank correlation between the model's F1 and the larger of the two overlap shares is 0.78 (0.44 for OpenAI Privacy Filter). This is consistent with, but does not prove, the interpretation that a model trained on the mix scores well on benchmarks that share its surface forms and poorly on text that does not (Section~\ref{sec:realtext}). Appendix~\ref{app:audit} lists the overlap per set.

\subsection{Generators, languages and evaluation overlap}
\label{sub:generators}

\paragraph{Generator attribution.} Table~\ref{tab:generators} lists the model strings recorded in the refined collection (317,846 rows; 226,546 of them are in the training mix). Sixteen distinct strings occur, plus 3.6\% of rows with none. About 21\% of rows are attributed to models outside the nine-model pool of Section~\ref{sec:pipeline} (qwen3-32b 7.4\%, gpt-5.4-mini 6.0\%, gemma-4-31b 4.0\%, MiMo-V2.5-Pro 3.6\%); Gemini 3.5 Flash accounts for 0.1\%. Repair rate is the share of rows flagged \texttt{had\_label\_repairs}; bad-span rate is the share of spans that are markup-contaminated or over-long (Section~\ref{sec:audit}). The rates are not comparable across models without controlling for language and scenario.

\begin{table}[!htbp]
\centering
\small
\caption{Refined collection by recorded generator (largest twelve of seventeen strings).}
\label{tab:generators}
\resizebox{\columnwidth}{!}{%
\begin{tabular}{l r r r r}
\toprule
Model string & Rows & Share (\%) & Repaired (\%) & Bad spans (\%) \\
\midrule
\texttt{mistral-medium-2505} & 145,873 & 45.9 & 34.8 & 4.7 \\
\texttt{gpt-oss-120b} & 55,652 & 17.5 & 41.7 & 5.6 \\
\texttt{qwen/qwen3-32b} & 23,447 & 7.4 & 71.9 & 8.0 \\
\texttt{glm-4.5-flash} & 22,258 & 7.0 & 34.2 & 4.4 \\
\texttt{gpt-5.4-mini} & 19,147 & 6.0 & 14.5 & 1.0 \\
\texttt{google/gemma-4-31b-it:free} & 12,694 & 4.0 & 26.1 & 5.1 \\
\texttt{<EMPTY>} & 11,513 & 3.6 & 0.0 & 9.1 \\
\texttt{mimo-v2.5-pro} & 11,311 & 3.6 & 0.0 & 1.4 \\
\texttt{\@cf/openai/gpt-oss-120b} & 7,896 & 2.5 & 40.6 & 5.2 \\
\texttt{deepseek-v4-flash-free} & 3,369 & 1.1 & 13.4 & 8.3 \\
\texttt{meta/llama-3.3-70b-instruct} & 2,709 & 0.9 & 33.9 & 3.2 \\
\texttt{qwen3-235b} & 793 & 0.2 & 45.6 & 6.9 \\

\bottomrule
\end{tabular}}
\end{table}

\begin{table}[!htbp]
\centering
\small
\caption{Refined collection by language. Markup: documents with a markup-contaminated span; 555 and Seq.: share of phone spans; Plc.: canonical placeholder names as a share of name spans; Bare: documents with a tagged value left bare elsewhere; Uniq.: unique-name ratio.}
\label{tab:languages}
\resizebox{\columnwidth}{!}{%
\begin{tabular}{l r r r r r r r}
\toprule
Language & Rows & Markup & 555 & Seq. & Plc. & Bare & Uniq. \\
\midrule
burmese & 20,508 & 20.9 & 0.0 & 1.4 & 0.0 & 57.2 & 0.26 \\
chinese & 19,154 & 21.1 & 0.1 & 3.2 & 16.1 & 34.7 & 0.10 \\
english & 27,757 & 29.4 & 58.4 & 10.5 & 5.1 & 40.4 & 0.28 \\
filipino & 16,515 & 27.2 & 0.4 & 5.7 & 8.9 & 43.2 & 0.22 \\
french & 15,459 & 36.3 & 0.1 & 9.5 & 0.0 & 38.2 & 0.24 \\
german & 16,490 & 31.1 & 0.1 & 25.7 & 1.0 & 38.6 & 0.26 \\
indonesian & 16,512 & 24.2 & 0.6 & 36.2 & 0.0 & 42.3 & 0.19 \\
japanese & 15,124 & 26.0 & 0.4 & 30.7 & 10.6 & 46.8 & 0.12 \\
korean & 15,718 & 24.2 & 0.3 & 38.0 & 9.4 & 33.4 & 0.09 \\
lao & 16,578 & 20.7 & 10.4 & 6.6 & 0.0 & 42.8 & 0.55 \\
malay & 16,303 & 24.6 & 0.2 & 11.6 & 0.0 & 40.5 & 0.19 \\
portuguese & 14,915 & 26.3 & 0.1 & 17.7 & 0.0 & 31.2 & 0.19 \\
russian & 17,260 & 20.2 & 0.3 & 14.5 & 0.6 & 30.5 & 0.15 \\
spanish & 18,451 & 28.7 & 0.7 & 12.5 & 0.0 & 35.9 & 0.30 \\
tamil & 16,121 & 27.8 & 0.1 & 11.9 & 0.0 & 47.8 & 0.22 \\
thai & 16,018 & 23.2 & 0.6 & 13.0 & 0.0 & 34.0 & 0.48 \\
vietnamese & 38,963 & 21.7 & 0.1 & 26.5 & 11.7 & 42.3 & 0.10 \\

\bottomrule
\end{tabular}}
\end{table}

\paragraph{Evaluation-set overlap.} Table~\ref{tab:overlap} gives, for each evaluation set, the share of gold spans whose exact (label, value) pair also occurs among training spans of self-generated rows and of rows of the same public source. Document-level overlap is zero everywhere.

\begin{table}[!htbp]
\centering
\small
\caption{Entity-value overlap between evaluation gold spans and the training mix (\% of gold spans).}
\label{tab:overlap}
\resizebox{\columnwidth}{!}{%
\begin{tabular}{l r r r}
\toprule
Evaluation set & Spans & generated rows & Same source \\
\midrule
ai4privacy\_de & 122,681 & 10.1 & 12.4 \\
ai4privacy\_en & 231,761 & 6.7 & 31.7 \\
ai4privacy\_es & 94,407 & 7.2 & 10.6 \\
ai4privacy\_fil & 30,011 & 15.6 & 16.9 \\
ai4privacy\_fr & 139,404 & 7.1 & 15.5 \\
ai4privacy\_id & 30,844 & 12.9 & 13.7 \\
ai4privacy\_ja & 29,733 & 3.7 & 5.7 \\
ai4privacy\_ko & 28,166 & 10.7 & 8.1 \\
ai4privacy\_ms & 29,749 & 14.4 & 14.1 \\
ai4privacy\_pt & 46,186 & 11.2 & 13.8 \\
ai4privacy\_vi & 30,408 & 11.8 & 50.7 \\
ai4privacy\_zh & 27,333 & 9.3 & 11.7 \\
creddata\_en & 2,204 & 0.0 & 0.0 \\
gretel\_en & 21,355 & 9.6 & 15.4 \\
nemotron\_en & 637,280 & 30.8 & 41.3 \\
eval-challenge & 69,823 & 28.7 & -- \\
eval-v2 & 9,249 & 16.1 & -- \\
eval & 29,919 & 55.6 & -- \\

\bottomrule
\end{tabular}}
\end{table}

\section{Human-Annotated Corpora: Results, Definitions and Alignment}
\label{app:realtext}

\subsection{How the three annotation schemes differ}
\label{sub:schemes}

Table~\ref{tab:defs} contrasts our definitions with what the two corpora annotate, as observed in their annotations and in our label maps (Table~\ref{tab:realmap}). The differences concern \emph{what counts} as PII (dates, locations, family terms, titles) and \emph{how long} a span is; they are independent of detector quality. One difference is not of this kind: hospital and institution names are inside our own \texttt{company\_name} definition, so a model trained on it should find them.

\begin{table*}[!htbp]
\centering
\small
\caption{Definition comparison. ``Observed'' entries describe the gold annotations we scored, not the full guideline documents.}
\label{tab:defs}
\resizebox{\textwidth}{!}{%
\begin{tabular}{p{0.13\textwidth} p{0.27\textwidth} p{0.27\textwidth} p{0.27\textwidth}}
\toprule
Concept & Ours (this work) & GraSCCo\_PHI (observed) & MEDDOCAN (observed) \\
\midrule
Person name & Patient, clinician and other person names; a role or title alone is not a name. The trained model often includes honorifics in the span. & Name only; honorific is a separate \textsc{name\_title} annotation (ignored here). & Name only; ``Dr.'', ``Dra.'' excluded. \\
Date & Specific dates tied to a person or encounter. & Every date, including a bare year (e.g.\ 2007). & Every temporal expression, including ``año 2004'' and durations (``3 años''). \\
Location & Street address, postal code, coordinates, care location. & Street, ZIP, city, \emph{country}. & Street, territory (city, province, region) and \emph{country}. \\
Organization & Hospitals, clinics, departments, companies. & Hospital and other organizations (\textsc{location\_hospital}, \textsc{\_organization}). & Hospital, health centre, institution. \\
Family terms & Only names are tagged. & Relative names only. & \textsc{familiares}: kinship words such as ``madre'' are annotated. \\
Age, sex, profession & Never tagged. & \textsc{age}, \textsc{profession} annotated (ignored here). & Age, sex, profession annotated (ignored here). \\
Identifiers & One label, \texttt{id\_number}. & One \textsc{id} label. & Five ID classes (insurance, contact, patient, licence, employment). \\
\bottomrule
\end{tabular}}
\end{table*}

\subsection{Results on the two corpora}
\label{sub:results2}

To test transfer beyond LLM-generated text we evaluate on two publicly shareable corpora annotated by humans: GraSCCo\_PHI \cite{lohr2024deidentifying} (63 German clinical documents derived from real documents by multiple alienation steps, 1,273 annotations that map to our ontology) and the MEDDOCAN test set \cite{marimon2019meddocan} (250 Spanish clinical case reports with surrogate PHI, 4,666 mapped annotations). Both are CC BY 4.0 and neither is a corpus of real patient records. Gold annotations whose source label has no counterpart in our ontology (age, profession, sex, honorific title) are ignored (Appendix~\ref{app:mappings}). We ran every system unchanged with the evaluator's pinned adapters; none of the text was used for training or tuning.

\begin{table}[!htbp]
\centering
\small
\caption{Exact and relaxed micro-F1 on human-annotated clinical-style text, with 95\% bootstrap intervals for exact F1. Best exact score in bold.}
\label{tab:realtext}
\resizebox{\columnwidth}{!}{%
\begin{tabular}{l ll ll}
\toprule
 & \multicolumn{2}{c}{GraSCCo\_PHI (de)} & \multicolumn{2}{c}{MEDDOCAN (es)} \\
System & Exact [95\% CI] & Rel. & Exact [95\% CI] & Rel. \\
\midrule
Ours & 0.583 [0.547, 0.619] & 0.787 & 0.720 [0.709, 0.730] & 0.817 \\
SuperClinical-Large & \textbf{0.619} [0.569, 0.663] & 0.797 & 0.589 [0.578, 0.601] & 0.821 \\
GLiNER2 & 0.419 [0.387, 0.454] & 0.544 & \textbf{0.745} [0.734, 0.756] & 0.798 \\
OpenAI Privacy Filter & 0.497 [0.455, 0.540] & 0.680 & 0.607 [0.595, 0.619] & 0.713 \\
LFM2.5-Encoder PII & 0.203 [0.173, 0.237] & 0.272 & 0.490 [0.478, 0.502] & 0.560 \\

\bottomrule
\end{tabular}}
\end{table}

\paragraph{Results.} Without any adaptation to these corpora, the model reaches exact F1 of 0.583 and 0.720 (relaxed 0.787 and 0.817), roughly three times its starting point on GraSCCo\_PHI (0.203) and 1.5 times on MEDDOCAN (0.490). It is second under each corpus's own conventions, behind SuperClinical-Large on GraSCCo\_PHI (overlapping intervals) and GLiNER2 on MEDDOCAN (Table~\ref{tab:realtext}).

\paragraph{Per-label comparison.} Table~\ref{tab:realtext-labels} breaks the comparison down by label. On GraSCCo\_PHI the model has the highest exact F1 on four of the seven labels (address 0.52, date 0.76, ID 0.64, phone 0.90) and ties for first on email (1.00, shared with GLiNER2, OpenAI Privacy Filter and LFM2.5-Encoder PII); on MEDDOCAN it is first on date (0.95) and within 0.02 of the best on email (0.97 against 0.99). Its lowest exact score is for person names on GraSCCo\_PHI (0.27), where its relaxed score of 0.87 is the highest of the five systems: the gap comes from the honorifics discussed below, not from missed names. No system detects hospital and institution names (exact F1 of at most 0.01 on both corpora).

\begin{table*}[!htbp]
\centering
\small
\caption{Exact / relaxed F1 per label on the two human-annotated corpora. ``--'' means the label does not occur in the gold annotation or was never scored. Hospital and institution names (\texttt{company\_name}) are annotated in both corpora; no system detects them.}
\label{tab:realtext-labels}
\resizebox{\textwidth}{!}{%
\begin{tabular}{l *{7}{c}}
\toprule
System & address & company & date & email & name & id & phone \\
\midrule
\multicolumn{8}{l}{\textit{GraSCCo\_PHI (de): exact / relaxed F1}} \\
Ours & 0.52/0.72 & 0.00/0.00 & 0.76/0.79 & 1.00/1.00 & 0.27/0.87 & 0.64/0.64 & 0.90/0.98 \\
SuperClinical-Large & 0.30/0.66 & 0.00/0.00 & 0.73/0.86 & 0.00/0.40 & 0.66/0.84 & 0.08/0.25 & 0.34/0.72 \\
GLiNER2 & 0.34/0.62 & 0.00/0.00 & 0.40/0.40 & 1.00/1.00 & 0.49/0.78 & 0.19/0.19 & 0.88/0.92 \\
OpenAI Privacy Filter & 0.06/0.36 & 0.00/0.00 & 0.73/0.78 & 1.00/1.00 & 0.19/0.63 & 0.39/0.39 & 0.72/0.94 \\
LFM2.5-Encoder PII & 0.09/0.30 & 0.00/0.14 & 0.15/0.15 & 1.00/1.00 & 0.35/0.47 & 0.36/0.36 & 0.00/0.29 \\
gold support & 135 & 38 & 694 & 1 & 322 & 58 & 25 \\
\midrule
\multicolumn{8}{l}{\textit{MEDDOCAN test (es): exact / relaxed F1}} \\
Ours & 0.58/0.73 & 0.00/0.00 & 0.95/0.97 & 0.97/0.99 & 0.81/0.96 & 0.78/0.82 & 0.22/0.24 \\
SuperClinical-Large & 0.48/0.81 & 0.01/0.05 & 0.94/0.96 & 0.91/0.97 & 0.78/0.89 & 0.12/0.66 & 0.84/0.87 \\
GLiNER2 & 0.75/0.84 & 0.00/0.00 & 0.15/0.15 & 0.96/0.99 & 0.89/0.93 & 0.87/0.88 & 0.55/0.59 \\
OpenAI Privacy Filter & 0.36/0.53 & 0.00/0.00 & 0.89/0.89 & 0.96/0.99 & 0.71/0.89 & 0.71/0.74 & 0.18/0.20 \\
LFM2.5-Encoder PII & 0.38/0.44 & 0.01/0.04 & 0.59/0.59 & 0.99/0.99 & 0.52/0.55 & 0.51/0.72 & 0.04/0.07 \\
gold support & 1732 & 203 & 611 & 249 & 1084 & 754 & 33 \\

\bottomrule
\end{tabular}}
\end{table*}

\paragraph{Error profile.} Table~\ref{tab:taxonomy} shows what happens to each gold span. Wrong-label predictions are rare (0\% of gold spans on GraSCCo\_PHI, 5\% on MEDDOCAN); most of the remaining error is span boundaries (20\% and 14\%) or spans with no overlapping prediction (31\% and 17\%), the latter dominated by the institution names that no system finds.

\begin{table}[!htbp]
\centering
\small
\caption{Fate of gold spans for ours (\% of scored gold spans): exact match, same label with different boundaries, other label only, no overlapping prediction.}
\label{tab:taxonomy}
\resizebox{\columnwidth}{!}{%
\begin{tabular}{l r r r r}
\toprule
Corpus & Exact & Boundary & Label & Missed \\
\midrule
GraSCCo\_PHI (de) & 48 & 20 & 0 & 31 \\
MEDDOCAN (es) & 64 & 14 & 5 & 17 \\
\bottomrule
\end{tabular}}
\end{table}

\subsection{Alignment rules}
\label{sub:align}

The two rules behind Table~\ref{tab:ladder} (Section~\ref{sec:realtext}) are applied identically to every system. Rule 1 strips a leading honorific (Herr, Frau, Dr., Prof., Sr., Sra., D\~{n}a., \ldots) from predicted names, since both corpora annotate the name only. Rule 2 ignores gold spans that our definition does not cover: countries, kinship words and non-specific dates (a bare year or a duration); 68 gold spans in GraSCCo\_PHI and 524 in MEDDOCAN. Applying rule 1 alone gives 0.683 for the model on GraSCCo\_PHI and 0.730 on MEDDOCAN. Hospital and institution names are not ignored in the table, because they are inside our definition; ignoring them as well (gold and predicted) would raise the model to 0.715 and 0.791, which shows only what the missing hospital detections cost. The rules were written after inspecting the errors, so the table is exploratory evidence, not a pre-registered result.

All predictions were produced by the evaluator's pinned adapters on one A10G GPU; the character offsets and labels of every prediction, and the scoring scripts, are included in the evaluation code. The corpora themselves are not redistributed.

\section{Training Recipe}
\label{app:training}

Table~\ref{tab:training} lists the settings of the single training run of our model described in Section~\ref{sec:model}. Hardware and wall-clock time were not recorded. The initialization is LFM2.5-Encoder-350M-PII-Detector, which is also one of our four baselines, so that baseline row is also the same network before training on our mix. Warm inference on one A10G GPU (model loading excluded; indicative, since each system uses its own adapter) takes 16.2 s for the 250 MEDDOCAN test documents (65 ms per document), within the 10.4--18.4 s of the four baselines.

\begin{table}[!htbp]
\centering
\small
\caption{Training recipe of our model (single run).}
\label{tab:training}
\resizebox{\columnwidth}{!}{%
\begin{tabular}{p{0.34\columnwidth} p{0.58\columnwidth}}
\toprule
Setting & Value \\
\midrule
Initialization & LFM2.5-Encoder-350M-PII-Detector \\
Adaptation & LoRA, rank 128, $\alpha=256$, dropout 0 \\
Head & Dropout 0.1 + Linear(1024, 37) \\
Decoder & Viterbi over valid BIOES transitions \\
Sequence length & 8,192 tokens; sequential packing with document isolation \\
Training data & 120,036 packed units from the training mix \\
Batch size & 128 packed units (global) \\
Optimizer & AdamW, lr $4\times10^{-4}$, constant, no warm-up, weight decay 0.01 \\
Precision & BF16 \\
Seed; checkpoint & 3407; optimizer step 250 of one continuous run \\
Selection & pinned control set, then the full benchmark \\
Splits & training rows are disjoint from every evaluation set (Section~\ref{sec:audit}); no separate validation set is described \\
Hardware & two GPUs, then four (per-device batch 64, then 32); GPU type and wall-clock time not recorded \\
\bottomrule
\end{tabular}}
\end{table}

\section{Benchmark Profile}
\label{app:benchmark}

Details behind the benchmark columns of Table~\ref{tab:corpora}. The benchmark has 5,100 documents (99,742 labelled spans): 1,700 standard and 3,400 adversarial documents, 300 per language in each of 17 languages.

\paragraph{Label distribution.} Table~\ref{tab:bench-labels} shows that each of the nine labels has between 7,256 and 15,993 spans (2.2:1) and occurs in at least 91.6\% of the documents; 4,446 documents (87.2\%) contain all nine. For comparison, the gold annotations of MEDDOCAN and GraSCCo\_PHI have 33 and 1 spans for their rarest label (52:1 and 694:1 against the most frequent; Table~\ref{tab:realtext-labels}), and neither annotates private URLs or secrets.

\begin{table}[!htbp]
\centering
\small
\caption{Label distribution of the benchmark.}
\label{tab:bench-labels}
\resizebox{\columnwidth}{!}{\begin{tabular}{l r r r}
\toprule
Label & Spans & Share (\%) & Docs with label (\%) \\
\midrule
\texttt{address} & 7,256 & 7.3 & 98.0 \\
\texttt{company\_name} & 9,181 & 9.2 & 95.8 \\
\texttt{date} & 10,857 & 10.9 & 91.6 \\
\texttt{email\_address} & 9,998 & 10.0 & 99.5 \\
\texttt{human\_name} & 15,806 & 15.8 & 99.9 \\
\texttt{id\_number} & 15,993 & 16.0 & 100.0 \\
\texttt{phone\_number} & 11,856 & 11.9 & 99.5 \\
\texttt{private\_url} & 8,381 & 8.4 & 99.7 \\
\texttt{secret} & 10,414 & 10.4 & 99.8 \\
\bottomrule
\end{tabular}}
\end{table}

\paragraph{Format diversity.} Table~\ref{tab:bench-formats} lists the ten text formats that each cover between 6.9\% and 14.0\% of the documents (4,985 documents); 74 further documents use ten other formats (under 1\% each, mostly general-domain and code-log text) and 41 documents have no format recorded. The two human-annotated corpora are each one language and one genre (clinical case reports, clinical documents).

\begin{table}[!htbp]
\centering
\small
\caption{Text formats of the benchmark (ten most frequent).}
\label{tab:bench-formats}
\resizebox{\columnwidth}{!}{\begin{tabular}{l r r}
\toprule
Text format & Documents & Share (\%) \\
\midrule
Voice-scribe transcript & 712 & 14.0 \\
Markdown sections & 573 & 11.2 \\
HL7-like pipe-delimited text & 536 & 10.5 \\
Patient portal audit log & 502 & 9.8 \\
JSON-like object & 487 & 9.5 \\
OCR text with light noise & 482 & 9.5 \\
FHIR-like JSON & 453 & 8.9 \\
Table-like rows & 451 & 8.8 \\
Messy nurse/admin note & 436 & 8.5 \\
Plain clinical note & 353 & 6.9 \\
\bottomrule
\end{tabular}}
\end{table}

\paragraph{Controlled variables.} Exactly 300 documents per language (100 standard, 200 adversarial); every adversarial document combines four or five of the thirteen perturbations of Table~\ref{tab:perturbations}; for every language the share of each label stays within 2.5 percentage points of its overall share, so comparisons across languages or conditions are not confounded by label mix.

\paragraph{What is not controlled.} The generator (3,508 documents, 68.8\%, by GPT-5.4-mini, which is not one of the nine training generators; 1,082 by Mistral Medium; the rest by smaller groups) and the scenario (1,611 of the 1,700 standard documents, 94.8\%, come from one scenario; the adversarial half is mostly two scenarios, adversarial obfuscation and all-labels-dense). The benchmark is not hand-adjudicated and shares an annotator with the training data (Section~\ref{sec:indomain}, Section~\ref{sec:limitations}).

\section{Public Benchmarks: Confidence Intervals}
\label{app:ci}

We recomputed the scores of Table~\ref{tab:benchmark-results} for all five systems from per-document predictions on the full evaluation splits of the 15 public sets (258,685 documents; pinned dataset revision, pinned adapters, one Modal A10G run per system), with the exact-match scorer of Section~\ref{sec:eval}. Table~\ref{tab:ci} gives 95\% percentile bootstrap intervals over documents (1,000 resamples, seed 0, the same resampled documents for all systems, so the paired difference is valid) for our model, for the strongest baseline on each set and for their difference. The difference is significant on every set except CredData ($^{ns}$): there the interval of the difference includes zero. On CredData our advantage over OpenAI Privacy Filter is not significant (0.590 vs.~0.584; difference $+$0.006, 95\% interval [$-$0.023, $+$0.035]).

\begin{table*}[!htbp]
\centering
\small
\caption{Exact-match micro-F1 with 95\% bootstrap intervals on the 15 public sets: ours, the strongest baseline on that set (OpenAI PF = OpenAI Privacy Filter) and the paired difference. $^{ns}$ = interval includes zero.}
\label{tab:ci}
\begin{tabular}{l r l l l l}
\toprule
Set & Docs & Ours [95\% CI] & Best baseline & Best baseline [95\% CI] & Difference [95\% CI] \\
\midrule
OpenPII (de) & 23,616 & 0.827 [0.824, 0.829] & OpenAI PF & 0.716 [0.713, 0.718] & +0.111 [+0.109, +0.113] \\
OpenPII (en) & 40,580 & 0.878 [0.876, 0.879] & OpenAI PF & 0.745 [0.743, 0.747] & +0.132 [+0.131, +0.134] \\
OpenPII (es) & 17,877 & 0.869 [0.867, 0.872] & OpenAI PF & 0.794 [0.791, 0.797] & +0.075 [+0.073, +0.078] \\
OpenPII (fil) & 5,087 & 0.848 [0.843, 0.852] & GLiNER2 & 0.690 [0.685, 0.696] & +0.158 [+0.153, +0.162] \\
OpenPII (fr) & 25,690 & 0.843 [0.841, 0.845] & OpenAI PF & 0.738 [0.735, 0.740] & +0.106 [+0.104, +0.108] \\
OpenPII (id) & 5,363 & 0.848 [0.844, 0.853] & GLiNER2 & 0.688 [0.682, 0.693] & +0.161 [+0.156, +0.165] \\
OpenPII (ja) & 5,027 & 0.751 [0.746, 0.756] & OpenAI PF & 0.549 [0.543, 0.555] & +0.202 [+0.196, +0.208] \\
OpenPII (ko) & 5,230 & 0.740 [0.734, 0.745] & OpenAI PF & 0.554 [0.548, 0.561] & +0.185 [+0.180, +0.192] \\
OpenPII (ms) & 5,089 & 0.839 [0.835, 0.844] & GLiNER2 & 0.692 [0.687, 0.697] & +0.147 [+0.142, +0.152] \\
OpenPII (pt) & 8,665 & 0.855 [0.851, 0.859] & OpenAI PF & 0.756 [0.752, 0.760] & +0.098 [+0.094, +0.102] \\
OpenPII (vi) & 5,195 & 0.880 [0.876, 0.884] & OpenAI PF & 0.634 [0.628, 0.640] & +0.245 [+0.240, +0.251] \\
OpenPII (zh) & 4,971 & 0.816 [0.811, 0.822] & OpenAI PF & 0.572 [0.565, 0.579] & +0.244 [+0.238, +0.251] \\
CredData (en) & 1,403 & 0.590 [0.567, 0.613] & OpenAI PF & 0.584 [0.559, 0.612] & +0.006 [-0.023, +0.035]$^{ns}$ \\
Gretel (en) & 5,000 & 0.872 [0.867, 0.878] & LFM2.5 & 0.729 [0.723, 0.736] & +0.143 [+0.135, +0.152] \\
Nemotron (en) & 99,892 & 0.956 [0.955, 0.956] & SuperClinical-L & 0.894 [0.893, 0.894] & +0.062 [+0.061, +0.063] \\
\midrule
Mean, all 15 & -- & 0.827 [0.826, 0.829] & OpenAI PF & 0.672 [0.670, 0.674] & +0.156 \\
\bottomrule
\end{tabular}
\end{table*}

\paragraph{Per-label scores.} Table~\ref{tab:perlabel} pools the 15 public sets and scores each label separately (a system is scored only on labels its fold can emit). Our model is best on every label, but it is weak on \texttt{address} (0.675) and, above all, \texttt{private\_url} (0.084): our ontology tags only access-bearing links and leaves public URLs untagged, while the public benchmarks tag every URL, so this label mostly measures a convention mismatch rather than detection quality.

\begin{table}[!htbp]
\centering
\small
\caption{Exact-match F1 per label, pooled over the 15 public benchmarks. ``--'' = the system's label fold cannot emit the label. Best per row in bold.}
\label{tab:perlabel}
\resizebox{\columnwidth}{!}{%
\begin{tabular}{l r r r r r}
\toprule
Label & Ours & OpenAI PF & GLiNER2 & SuperClinical-L & LFM2.5 \\
\midrule
\texttt{address} & \textbf{0.675} & 0.261 & 0.486 & 0.534 & 0.454 \\
\texttt{company\_name} & \textbf{0.952} & -- & -- & 0.878 & 0.918 \\
\texttt{date} & \textbf{0.963} & 0.896 & 0.670 & 0.885 & 0.492 \\
\texttt{email\_address} & \textbf{0.988} & 0.949 & 0.946 & 0.918 & 0.952 \\
\texttt{human\_name} & \textbf{0.918} & 0.763 & 0.591 & 0.694 & 0.508 \\
\texttt{id\_number} & \textbf{0.957} & 0.847 & 0.737 & 0.774 & 0.599 \\
\texttt{phone\_number} & \textbf{0.930} & 0.827 & 0.867 & 0.768 & 0.788 \\
\texttt{private\_url} & \textbf{0.084} & 0.001 & -- & 0.001 & 0.000 \\
\texttt{secret} & \textbf{0.918} & 0.539 & 0.597 & 0.693 & 0.317 \\
\bottomrule
\end{tabular}}
\end{table}

\section{Benchmark Breakdown}
\label{app:benchdeep}

\paragraph{Reading the in-domain result.} The benchmark is produced by the same pipeline and annotation procedure as the training data, so we use it as a regression test of the pipeline and do not interpret the margin as evidence of superior handling of the ontology or perturbations; a baseline that is penalized for tags the reference omits is not comparable to a model trained on the same conventions. Three breakdowns (below) qualify this. \emph{Format:} the margin over the best baseline holds in every one of the 11 text formats; our model is weakest on FHIR-like JSON (0.738) and HL7-like messages (0.828), where the baselines are also weakest (OpenAI Privacy Filter 0.458 and 0.424), and strongest on Markdown (0.954) and table-like text (0.936). \emph{Generator:} 69\% of the benchmark was written by GPT-5.4-mini, which is not a training generator, and our model scores 0.893 on those documents against 0.859 on the rest; within the standard split alone it scores 0.923 [0.906, 0.940] on the 108 held-out-generator documents and 0.859 [0.850, 0.867] on the others. All systems score higher on GPT-5.4-mini text, so these documents are easier, not specially favourable to our model, and we see no sign that it depends on its training generators (the contrast is partly confounded with split and sample size). \emph{Labels:} the baselines lose recall (e.g.~0.33 for secrets) as well as precision (0.47 for names), so the gap is not only a convention effect. We have not hand-adjudicated the benchmark.

Exact-match micro-F1 of the five systems on our benchmark (5,100 documents), recomputed from per-document predictions (one Modal run per system, same adapters and label folds as Section~\ref{sec:eval}). The score in Table~\ref{tab:benchmark-results} is the unweighted mean of the standard and the adversarial split (0.878 for ours); pooled over all documents it is 0.884. Intervals are 95\% bootstrap intervals over documents (1,000 resamples) for our model. Table~\ref{tab:bench-format} breaks the scores down by text format, Table~\ref{tab:bench-gen} by split and by generator.

\begin{table*}[!htbp]
\centering
\small
\caption{Exact-match F1 on our benchmark by text format. Best per row in bold.}
\label{tab:bench-format}
\begin{tabular}{l r l r r r r}
\toprule
Group & Docs & Ours [95\% CI] & OpenAI PF & GLiNER2 & SuperClinical-L & LFM2.5 \\
\midrule
voice-scribe transcript & 712 & \textbf{0.910 [0.902, 0.917]} & 0.559 & 0.454 & 0.257 & 0.279 \\
Markdown sections & 573 & \textbf{0.954 [0.950, 0.959]} & 0.617 & 0.508 & 0.271 & 0.310 \\
HL7-like pipe-delimited text & 536 & \textbf{0.828 [0.818, 0.837]} & 0.424 & 0.429 & 0.180 & 0.216 \\
patient portal audit log & 502 & \textbf{0.887 [0.874, 0.899]} & 0.528 & 0.458 & 0.226 & 0.277 \\
JSON-like object & 487 & \textbf{0.892 [0.882, 0.902]} & 0.602 & 0.537 & 0.243 & 0.194 \\
OCR text with light noise & 482 & \textbf{0.904 [0.894, 0.912]} & 0.577 & 0.489 & 0.243 & 0.288 \\
FHIR-like JSON & 453 & \textbf{0.738 [0.725, 0.751]} & 0.458 & 0.417 & 0.216 & 0.147 \\
table-like rows & 451 & \textbf{0.936 [0.928, 0.945]} & 0.578 & 0.510 & 0.246 & 0.306 \\
messy nurse/admin note & 436 & \textbf{0.897 [0.887, 0.907]} & 0.580 & 0.494 & 0.256 & 0.321 \\
plain clinical note & 353 & \textbf{0.922 [0.914, 0.929]} & 0.594 & 0.477 & 0.249 & 0.325 \\
clinical dialogue transcript & 43 & \textbf{0.822 [0.768, 0.876]} & 0.490 & 0.409 & 0.296 & 0.268 \\
(no format recorded) & 41 & \textbf{0.834 [0.777, 0.883]} & 0.639 & 0.498 & 0.365 & 0.376 \\
\midrule
All documents (pooled) & 5,100 & \textbf{0.884 [0.880, 0.887]} & 0.548 & 0.474 & 0.238 & 0.265 \\
\bottomrule
\end{tabular}
\end{table*}

\begin{table*}[!htbp]
\centering
\small
\caption{Exact-match F1 on our benchmark by split and generator. Documents generated by GPT-5.4-mini are held out from training; all adversarial documents come from it. Best per row in bold.}
\label{tab:bench-gen}
\begin{tabular}{l r l r r r r}
\toprule
Group & Docs & Ours [95\% CI] & OpenAI PF & GLiNER2 & SuperClinical-L & LFM2.5 \\
\midrule
adversarial & 3,400 & \textbf{0.892 [0.889, 0.896]} & 0.558 & 0.473 & 0.228 & 0.260 \\
standard & 1,700 & \textbf{0.863 [0.856, 0.872]} & 0.524 & 0.478 & 0.260 & 0.277 \\
adversarial (all GPT-5.4-mini) & 3,400 & \textbf{0.892 [0.889, 0.896]} & 0.558 & 0.473 & 0.228 & 0.260 \\
standard: other generators & 1,592 & \textbf{0.859 [0.850, 0.867]} & 0.516 & 0.474 & 0.259 & 0.274 \\
standard: held-out (GPT-5.4-mini) & 108 & \textbf{0.923 [0.906, 0.940]} & 0.619 & 0.519 & 0.270 & 0.314 \\
\bottomrule
\end{tabular}
\end{table*}

\paragraph{Label recall and precision.} On the benchmark, OpenAI Privacy Filter has recall 0.33 for \texttt{secret}, 0.42 for \texttt{phone\_number} and 0.43 for \texttt{email\_address}, and precision of only 0.48 for \texttt{address} and 0.47 for \texttt{human\_name}; our model has 0.78--0.95 on both for every label. The baselines' gap is therefore partly missed identifiers and partly extra tags that the reference omits.

\section{MultiGraSCCo: All Ten Languages}
\label{app:multigrascco}

We use the MultiGraSCCo annotation release (Zenodo, CC BY 4.0), which contains the 63 GraSCCo documents in ten languages with a direct-identifier (PHI) layer and an indirect-identifier (IPI) layer. We score only the PHI layer, whose types follow GraSCCo\_PHI, with the label map of Appendix~\ref{app:mappings}. Documents whose released annotation offsets do not reproduce the annotated text were excluded (column Docs). Scoring, systems and pinned code are as in Section~\ref{sec:eval}, and the model is the released checkpoint at its pinned revision. German is the original GraSCCo\_PHI and reproduces Table~\ref{tab:realtext} for ours (0.583 exact F1); two baselines differ from that table by at most 0.003 (SuperClinical-Large 0.616 against 0.619, LFM2.5 0.205 against 0.203), which we did not investigate. The ten test sets are one corpus translated automatically, so they are not independent samples, and the analysis is exploratory.

\begin{table}[!htbp]
\centering
\small
\caption{Exact micro-F1 on all ten MultiGraSCCo languages (PHI layer). Docs = documents scored. $^\ast$ = language in our training data. Best per row in bold. 95\% bootstrap intervals for ours: German 0.547--0.621, English 0.638--0.708, French 0.685--0.744, Russian 0.671--0.753, Italian 0.652--0.720, Polish 0.611--0.684, Ukrainian 0.654--0.748, Turkish 0.595--0.674, Arabic 0.202--0.314, Persian 0.045--0.103.}
\label{tab:multigrascco-exact}
\resizebox{\columnwidth}{!}{%
\begin{tabular}{l r ccccc}
\toprule
Language & Docs & Ours & \shortstack{OpenAI\\PF} & GLiNER2 & \shortstack{SuperClinical\\Large} & \shortstack{LFM2.5\\PII} \\
\midrule
German$^\ast$ & 63 & 0.583 & 0.497 & 0.419 & \textbf{0.616} & 0.205 \\
English$^\ast$ & 60 & 0.674 & 0.477 & 0.518 & \textbf{0.776} & 0.233 \\
French$^\ast$ & 60 & 0.715 & 0.601 & 0.516 & \textbf{0.717} & 0.237 \\
Russian$^\ast$ & 60 & \textbf{0.713} & 0.538 & 0.502 & 0.422 & 0.201 \\
Italian & 60 & 0.686 & 0.644 & 0.512 & \textbf{0.692} & 0.205 \\
Polish & 60 & \textbf{0.647} & 0.524 & 0.451 & 0.564 & 0.206 \\
Ukrainian & 60 & \textbf{0.701} & 0.488 & 0.464 & 0.439 & 0.200 \\
Turkish & 60 & \textbf{0.638} & 0.413 & 0.234 & 0.528 & 0.201 \\
Arabic & 61 & 0.254 & 0.448 & \textbf{0.533} & 0.137 & 0.042 \\
Persian & 62 & 0.073 & 0.433 & \textbf{0.515} & 0.068 & 0.008 \\
\bottomrule
\end{tabular}}
\end{table}

\begin{table}[!htbp]
\centering
\small
\caption{Relaxed micro-F1 (same label, at least one overlapping character) on all ten MultiGraSCCo languages.}
\label{tab:multigrascco-relaxed}
\resizebox{\columnwidth}{!}{%
\begin{tabular}{l r ccccc}
\toprule
Language & Docs & Ours & \shortstack{OpenAI\\PF} & GLiNER2 & \shortstack{SuperClinical\\Large} & \shortstack{LFM2.5\\PII} \\
\midrule
German$^\ast$ & 63 & 0.787 & 0.679 & 0.545 & \textbf{0.796} & 0.272 \\
English$^\ast$ & 60 & 0.810 & 0.726 & 0.635 & \textbf{0.873} & 0.363 \\
French$^\ast$ & 60 & 0.823 & 0.752 & 0.635 & \textbf{0.838} & 0.275 \\
Russian$^\ast$ & 60 & \textbf{0.828} & 0.766 & 0.590 & 0.725 & 0.286 \\
Italian & 60 & 0.816 & 0.773 & 0.604 & \textbf{0.820} & 0.254 \\
Polish & 60 & \textbf{0.803} & 0.685 & 0.567 & 0.788 & 0.253 \\
Ukrainian & 60 & \textbf{0.801} & 0.710 & 0.576 & 0.726 & 0.278 \\
Turkish & 60 & \textbf{0.760} & 0.608 & 0.596 & 0.733 & 0.264 \\
Arabic & 61 & 0.428 & \textbf{0.657} & 0.619 & 0.425 & 0.115 \\
Persian & 62 & 0.122 & 0.622 & \textbf{0.633} & 0.383 & 0.019 \\
\bottomrule
\end{tabular}}
\end{table}

\paragraph{Where ours is not best.} Ours is best on exact F1 in Russian, Polish, Ukrainian and Turkish; SuperClinical-Large is best in German, English, French and Italian, mostly by small margins, and GLiNER2 is best in Arabic and Persian. In Arabic and Persian ours reaches 0.254 and 0.073 (OpenAI Privacy Filter 0.448 and 0.433). No Arabic-script language is in our training data. In both languages most dates, which are written with Arabic-Indic or Persian digits, are missed (overlap recall of date spans 0.19 and 0.00, against 0.67 in German), but we did not test whether the digits explain the failure. Averaged over the ten languages the exact F1 is 0.568 for ours, 0.506 for OpenAI Privacy Filter, 0.496 for SuperClinical-Large, 0.466 for GLiNER2 and 0.174 for LFM2.5 (unweighted means); on relaxed F1 ours is tied with OpenAI Privacy Filter (0.698) and behind SuperClinical-Large (0.711).

\end{document}